\pdfoutput=1
\documentclass[conference]{IEEEtran}

\usepackage{cite}
\usepackage{amsmath,amssymb,amsfonts}
\usepackage{algorithmic}
\usepackage{graphicx}
\usepackage{textcomp}
\usepackage{xcolor}
\usepackage{url}
\usepackage[T1]{fontenc}
\usepackage[utf8]{inputenc}

\def\BibTeX{{\rm B\kern-.05em{\sc i\kern-.025em b}\kern-.08em
    T\kern-.1667em\lower.7ex\hbox{E}\kern-.125emX}}
\usepackage{graphicx}

\begin{document}

\title{PULP-NN: Accelerating Quantized Neural Networks on Parallel Ultra-Low-Power RISC-V Processors}

\author{\IEEEauthorblockN{Angelo Garofalo\IEEEauthorrefmark{2}, Manuele Rusci\IEEEauthorrefmark{2}, Francesco Conti\IEEEauthorrefmark{2}\IEEEauthorrefmark{1}, Davide Rossi\IEEEauthorrefmark{2} and Luca Benini\IEEEauthorrefmark{2}\IEEEauthorrefmark{1}} 
\IEEEauthorblockA{\textit{DEI, University of Bologna, Italy}\IEEEauthorrefmark{2} \quad \textit{IIS lab, ETH Zurich, Switzerland}\IEEEauthorrefmark{1} \\
 $\{$angelo.garofalo, manuele.rusci, davide.rossi$\}$@unibo.it \qquad $\{$fconti, lbenini$\}$@iis.ee.ethz.ch}}

\maketitle

\begin{abstract}

We present PULP-NN, an optimized computing library for a parallel ultra-low-power tightly coupled cluster of RISC-V processors.
The key innovation in PULP-NN is a set of kernels for Quantized Neural Network (QNN) inference, targeting byte and sub-byte data types, down to INT-1, tuned for the recent trend toward aggressive quantization in deep neural network inference.
The proposed library exploits both the digital signal processing (DSP) extensions available in the PULP RISC-V processors and the cluster's parallelism, achieving up to $15.5$ MACs/cycle on INT-8 
%only $51\%$ below the theoretical peak performance for a software programmable cluster of parallel processors,
and improving performance by up to 63$\times$ with respect to a sequential implementation on a single RISC-V core implementing the baseline RV32IMC ISA.
Using PULP-NN, a CIFAR-10 network on an octa-core cluster runs in $30 \times$ and $19.6 \times$ less clock cycles than the current state-of-the-art ARM CMSIS-NN library, running on STM32L4 and STM32H7 MCUs, respectively.
The proposed library, when running on GAP-8 processor, outperforms by $36.8 \times$ and by $7.45 \times$ the execution on energy efficient MCUs such as STM32L4 and high-end MCUs such as STM32H7 respectively, when operating at the maximum frequency.
The energy efficiency on GAP-8 is $14.1 \times$ higher than STM32L4 and $39.5 \times$ higher than STM32H7, at the maximum efficiency operating point.
\end{abstract}

\section{Introduction}
\label{sec:Introduction}
The Internet-of-Things has favored a rapid growth of the number of wireless-connected nodes for a large variety of applications, including agriculture \cite{elijah2018overview}, health monitoring \cite{hassanalieragh2015health}, surveillance \cite{motlagh2017uav}, structural monitoring \cite{tokognon2017structural}.
Such a massive unconstrained increment poses severe challenges to the network infrastructure, due to the exponential increase of data flowing through the network.
Capacity, security and reliability issues are exacerbated as the number of IoT nodes increases exponentially together with the ability to produce high-bandwidth data.

To address IoT scalability issues, data must be filtered at the edge of the network, on the sensor system itself \cite{shi2016edge}, using compression and analytics algorithms.
To this aim, Machine Learning (ML), including also state-of-the-art Deep Learning (DL), provides attractive solutions for edge processing. ML algorithms ``squeeze'' raw sensor data in a much more semantically dense format (i.e. classes or extracted high-level features/symbols), eventually packed into few bytes of information for wireless transmission.
%of only few bytes that need to be wirelessly transmitted. 
%

To empower IoT nodes with smart capabilities \cite{conti2017iot}, the design process of edge devices must trade-off the high computation and memory requirements of leading DL methods with the usual scarcity of resources of deeply embedded systems, powered by batteries or energy harvesters. 
Typically, deep network inference tasks run on GPUs or FPGAs devices, which however have a power envelope significantly higher than what can be sustained on extreme-edge devices, integrated with the sensors.
On the other side of the spectrum, resource-constrained MCUs are flexible, due to their software programmability, low-cost, low-power and suitable for extreme-edge usage, but they present severe limitations in memory footprint and computation resources that may prevent meeting application-specific latency and accuracy requirements.

To reduce the computational cost and memory footprint of Neural Newtorks, so that they can fit the limited computing capability and storage capacity of MCU-class devices, recent progress in DL training methodologies has introduced novel quantization methods, aiming at compressing either network weights parameters or activations into 8-bit or smaller data types, while incurring into a reduced or even negligible accuracy loss \cite{hubara2017quantized,lin2016fixed,wang2018haq,jacob2018quantization, moons2017minimum, conti2018xnor, rusci2019memory}.
%
%Hence, the embedded processing
Since Quantized Neural Networks (QNNs) feature much lower memory requirements than 32-bit floating point full precision models and low-bitwidth fixed-point execution units can operate efficiently at the core of the convolution routine, industry and academia are devoting a major effort to develop hardware and software platforms for efficient execution of QNNs on MCU-class devices.

In this work, we propose the first multicore computing library for QNN inference on fully programmable edge devices, which supports low bit-width (8-bit, 4-bit, 2-bit and 1-bit) operations.
%We also study the best implementation solutions to address the deployment of the DL at the edge from a computing perspective. 
While efficient libraries for commercial MCUs have been proposed for edge QNN inference \cite{lai2018cmsis, rusci2018work}, not many software solutions have been yet presented that efficiently exploit a parallel MCU architecture. We fill this void by building the back-end library upon the recent architectural template of parallel ultra-low-power RISC-V based platforms such as GAP8 \cite{flamand2018gap}, which improve energy efficiency and performance in IoT edge devices coupling parallelism with low voltage operation \cite{rossi2017energy}.
%\cc{Addressing the computational issues of the DL deployment on MCUs, in this work we propose a significant step forward toward the QNN inference at the edge on fully programmable devices, building upon the recent architectural template of parallel ultra-low-power platforms such as GAP8 \cite{flamand2018gap}, which improves energy efficiency and performance in IoT edge devices coupling parallelism with low voltage operation.}
%\st{In this work, we propose a significant step forward toward software programmable Quantized Neural Networks inference on MCU-class programmable devices, building upon the recent architectural template of parallel ultra-low-power platforms such as GAP8} \cite{flamand2018gap} \st{, which improve energy efficiency and performance in IoT edge devices coupling parallelism with low voltage operation.}
%
%While efficient libraries for commercial MCUs have been proposed for edge QNN inference \cite{lai2018cmsis, rusci2018work}, not many software solutions have been yet presented that efficiently exploit a parallel MCU architecture: in this work, we target the goal to fill this void by proposing \st{an open-source} \cc{a back-end software} library targeting a Parallel Ultra-Low-Power cluster of RISC-V based processors.
%
The main contributions of this paper are the following:
\begin{itemize}
    \item \textit{PULP-NN} \footnote{https://github.com/pulp-platform/pulp-nn}, an open-source optimized library based on the CMSIS-NN~\cite{lai2018cmsis, rusci2018work} dataflow including a full set of kernels and utilities to support the inference of Quantized Neural Networks (8,4,2 and 1-bit) on a DSP-optimized RISC-V based processor. 
    By fully exploiting the DSP extensions available within the ISA, we can achieve a speedup of $ 9 \times$ with respect to  a plain\textit{RV32IMC} ISA;
    \item We optimized the library for a Parallel Ultra-Low-Power (PULP) cluster of RISC-V processors, leading to near-linear speedup with respect to single core execution, increasing the throughput of each kernel by up to $7.5 \times$ on eight cores; 
    \item We optimized the convolution kernel, the most computing intensive task of CNN workloads, by improving data reuse, with a further 20\% performance gain with respect to the original kernel of CMSIS-NN~\cite{lai2018cmsis}, with a $\sim$1.9$\times$ improvement with respect to the GAP-8 NN native library and an overall efficiency of $49\%$ in terms of MAC utilization, which implies just 1.01 LD/ST per MAC, and brings us to just a factor of 2 from the theoretical peak MAC utilization achievable using only register operands; 
    \item We compare our solution with State-of-the-Art architectures and software, by running a CIFAR-10 quantized model on the GAP8 8-core cluster, outperforming by $19.5 \times$ a high-end MCU (based on ARM CORTEX-M7) running the same network using the CMSIS-NN library. The inference with the proposed library also achieves $14.1 \times$ better energy efficiency with respect to a highly energy efficient MCU (based on ARM CORTEX-M4).
    % In this case, we are able to propose a speedup in the whole network execution of $19.49 \times$ with respect to the inference of the same network on an off-the-shelf microcontroller, with an Arm CORTEX-M7 core, using CMSIS-NN library.
\end{itemize}

\noindent These order-of-magnitude improvements with respect to State-of-the-Art MCUs demonstrate for the first time that extreme-edge inference of QNN models is indeed possible on today's parallel ultra-low power MCUs.

\section{Related Work}
\label{sec:Related_work}
The success of Deep Learning (DL) has paved the way to many different DL deployments on embedded computing platforms of all kinds.
In this section, we recap the state-of-the-art and give insights on its applicability to CNN inference at the extreme-edge, on IoT end-nodes.
%explain if/how it applies to CNN inference on IoT end-nodes.

% FPGA APPROACHES %
\subsubsection*{\textbf{FPGA Based Approaches}}
Recent heterogeneous FPGAs such as Xilinx Zynq have enabled many solutions for CNN acceleration, embedding general purpose processors that manage the program flow, handle I/O and memory accesses, making them easier to program.
As DSP-capable FPGAs have a power envelope in the order of Watts, numerical precision of the CNN operands plays a crucial role to achieve high performance and thus energy efficiency.
While several architectures available in literature feature a precision of 16-bit (fixed-point) \cite{gokhale2017snowflake,ma2017automatic,venieris2017latency, meloni2018neura}, more and more designs are moving towards lower precision.
For example, Qiu et al. \cite{qiu2016going} proposed a CNN accelerator supporting 8 and 4-bit data, implemented on a Xilinx Zynq platform.
On this trail, even extreme quantization approaches have been presented, exploiting ternary or binary networks \cite{prost2017scalable, umuroglu2017finn}.
While most DSP-capable FPGAs currently do not offer a low enough power envelope to be used in IoT end-nodes,
Lattice recently announced SenseAI class of FPGAs \cite{LatticeSENSEAI} providing a comprehensive hardware and software solutions for always-on artificial intelligence (AI) within a power budget between 1 mW and 1 W. However these ultra-low power FPGAs are currently too expensive for many applications where MCUs are traditionally chosen because of their low cost. Furthermore, they report \cite{senseaiwhite} a measured performance of 8 fps with 64$\times$64 RGB input for a VGG8 like 16-bit  CNN at a power consumption of 7 mW, which maps to $0.88\,mJ/frame$, and  performance of 5 fps for a VGG network consisting of 6 convolution layers and 4 fully connected at a power consumption of 3.3 mW with an energy per inference of $0.66\,mJ/frame$. Both the results are significantly higher ($4.63 \times$ and $3.48 \times$, respectively) than the energy per frame that we report at the maximum efficiency point for our solution in sec. \ref{sec:results}.

% APPLICATION SPECIFIC ARCHITECTURES %
\subsubsection*{\textbf{Application Specific Architectures}}
On the other side of the programmability spectrum, ASIC accelerators are known to achieve best in class performance and energy efficiency.
Notable examples are Orlando~\cite{desoli201714} achieving energy efficiencies in the order of a few Top/s/W, and Origami~\cite{cavigelli2017origami} achieving a throughput of 274 Gop/s, with an efficiency of 803 Gop/s/W.
Dropping the arithmetic precision of CNN operands has demonstrated to be a useful technique to reduce the memory footprint and the energy cost for computation \cite{zhou2016dorefa, courbariaux2015binaryconnect, courbariaux2016binarized, rastegari2016xnor}.
UNPU \cite{lee2018unpu} is an example of an accelerator targeting fully-variable weight bit-precision, achieving a peak energy efficiency of 50.6 Top/s/W at a throughput of 184 Gop/s.
YodaNN \cite{andri2018yodann} targets binary-weight networks and reaches energy efficiency up to 61 Top/s/W.
Other accelerators exploit extreme quantization for the deployment of binary neural networks on silicon using in- or near-memory computing techniques (e.g., Brein~\cite{ando2018brein}, Conv-RAM \cite{biswas2018conv}) with energy efficiencies in the range 20-55 Top/s/W.
Such high energy efficiency and throughput achievable using ASIC accelerators are counterbalanced by limited flexibility, being application specific, which makes them unattractive to satisfy fully the flexibility demand of IoT edge nodes.
\begin{table*}[t]
   \centering
  \begin{tabular}{ccccc}
    %\toprule
    \multicolumn{5}{c}{\textbf{Summary of CNN Embedded Inference Computing Platform}} \\ \\
     \hline
     & Performance & Energy Efficiency & Power Budget & Flexibility \\ \\

     \textbf{ASICs \cite{lee2018unpu, andri2018yodann, cavigelli2017origami, desoli201714}}  & 1 - 10 Tops/s & 10 - 100 Tops/s/W & 1 mW - 1 W &  Low\\ \\ 
     \textbf{FPGAs \cite{gokhale2017snowflake,ma2017automatic,venieris2017latency, meloni2018neura, qiu2016going}} & 10 - 200 Gops/s &  1 - 10 Gops/s/W & 1 W - 10 W &  Medium\\  \\

     %\textbf{GP-GPUs \cite{NvidiaTegra2015perf, NvidiaTegra2015, NvidiaTuring}} & 100 - 500 Tops/s & 0.5 - 5 Tops/s/W & 100 W - 300 W & Medium\\ 
     %\hline
     \textbf{MCUs \cite{STM32L476, STM32H743xl} } & 100 - 300 Mops/s & 1 - 3 Gops/s/W & 1 mW - 1 W & High\\  \\
     \textbf{PULP SoCs \cite{conti2017iot, flamand2018gap, pullini2019mr} } & 1 - 2 Gops/s & 30 - 50 Gops/s/W & 1 mW - 100 mW & High\\ \\
        
    \hline
  %\bottomrule
  
\end{tabular}
\newline
\caption{The table shows the trade-offs among the CNN computing platforms described in the related work section.}
 \label{tab:related_work}
\end{table*}
%%% SOFTWARE SOLUTIONS %%%
\subsubsection*{\textbf{Software Programmable Architectures}}
Software-programmable general-purpose processors provide the highest degree of flexibility in QNN inference at the edge.
While CNNs are traditionally executed on programmable high-performance GPUs \cite{NvidiaTegra2015, NvidiaTegra2015perf} also with reduced precision support ~\cite{NvidiaTuring}, these platforms are typically not designed to operate in the tight power envelope of IoT end-nodes, and their cost is off-spec too.
Some architectures exploit the computing power of multi-core processors, such as Raspberry Pi 3+ \cite{Raspberry}, powered by a Quad-core ARM CORTEX-A53. 
Although these platforms are relatively inexpensive and flexible, their power consumption is too high as well.

To fit the power budget of IoT edge devices, many low power microcontrollers include ARM CORTEX-M cores. Among these solutions, STMicroelectronics proposed low-end (STM32L4 family based on ARM CORTEX M-4 cores and high-end (STM32H7 family featuring ARM CORTEX M-7 cores) microcontrollers supporting DL processing at the edge \cite{STM32L476, STM32H743xl}.
To improve the computing capabilities of such tiny and cheap computing platforms, ARM recently announced the development of the ARMv8.1-M \cite{Armv8.1} architecture, featuring Helium, an ISA extension tailored for DSP-oriented workloads, such as an inference task. However, such an extension is not supported yet by any device.

Other solutions move toward heterogeneous architectures, coupling microcontrollers with dedicated CNN accelerators, to deal with the extremely regular CNN workload. ARM proposed Trilium \cite{trilium}, a heterogeneous compute platform which provides flexible support for ML workloads. Conti et al. \cite{conti2015ultra} proposed a convolution engine to be integrated in a microcontroller to speed up the convolutional kernels while Kendryte \cite{Kendryte} is a dual-core RISC-V SoC outfitted with a CNN accelerator for AI applications. Flamand~et~al. proposed GAP8~\cite{flamand2018gap}, a multi-GOPS fully programmable RISC-V IoT-edge computing engine, featuring a cluster of 8 cores with dedicated DSP extensions and a CNN-specialized accelerator. These accelerators can give the MCU a 5 to 10$\times$ energy efficiency boost, but they are proprietary, closed, platform specific and currently not fully supported by the software design flows. Hence, their acceptance and penetration among application developers is still quite low.

Table \ref{tab:related_work} summarizes the trade-offs among the CNN computing platforms described so far. Next section will describe the State-of-the-Art of software solutions for MCU platforms, the main focus of this work.

\subsubsection*{\textbf{Optimized Software Libraries}}
On the MCU side, the limited computational and memory capabilities make aggressive software and algorithmic optimizations necessary to deploy DNN inference models on them.
An efficient solution to reduce DNN memory footprint is to use fixed-point arithmetic and quantization of both weights and activations into 8-bit or smaller data types, at the cost of a minor drop in accuracy \cite{hubara2017quantized,lin2016fixed, rusci2019memory}.
%
%Choi et al. \cite{choi2018pact}, for example, have proved that the 4-bit quantized CNNs they consider achieve an accuracy similar to single-precision floating point representation.
%
%The accuracy drop is limited to $3 \%$ when applying ResNet50 topology on Imagenet dataset with 2-bit weights and 4-bit activations and to $6.5 \%$ when downscaling the weights and activations to 2 bits.
%
Relying on fixed-point quantized networks, ARM proposed the CMSIS-NN library \cite{lai2018cmsis}, which maximizes the performance of the DL kernels on CORTEX-M series cores, supporting 16-bit and 8-bit fixed-point data.
On the same trail, targeting a parallel MCU architecture such as GAP-8, Greenwaves Technologies released open-source a set of QNN kernels (16- and 8-bit data precisions) as part of a proprietary tiling solution \cite{flamand2018gap}. The tiling procedure, exploiting the DMA controller available on GAP-8, hides the latency of fetching/storing activations and weights along the memory hierarchy introducing only a small overhead (a few \%), thus enabling the processing of large networks whose layers may not fit the MCU on-board memory. In this work we focus on the computational aspects of reduced precision quantized CNN inference. In this context, despite the demonstrated effectiveness of sub-byte aggressive quantization \cite{moons2017minimum}, only Rusci~et~al.~\cite{rusci2018work} explored the inference speed as well as memory requirements of using low-precision (4-, 2- or 1-bit) convolution kernels on a Cortex-M7 microcontroller.

Our work aims at bridging this gap, leveraging the results of \cite{moons2017minimum} and focusing on the computational side to enable efficient QNN inference at the edge on fully programmable devices.
%We adopt CMSIS-NN as a baseline to measure the compelling benefits of the solutions we propose in this paper.
%
To this purpose, we propose an open-source QNN library targeting 8-bit as well as sub-byte quantized data types, down to 1-bit data, targeting parallel ultra-low-power (PULP) architectures.
By exploiting the ISA extensions available on PULP architectures and tightly coupled cluster, our contributions outperform the CMSIS-NN based solutions by one order of magnitude in terms of performance and energy efficiency.

%%%%BACKGROUND %%%
\section{Background}
\label{sec:Background}
\subsection{Quantized Neural Networks}
\label{sec:background_quant}
A Deep convolution Neural Network (CNN) is made of several layers stacked one on top of the other.
Each layer can be considered as a computation kernel, and the most computive-intensive ones are the convolution and the fully connected layers.

To favor the deployment of CNN models into resource-constrained devices, a set of constraints can be applied to the numeric domain of either network parameters or activation values, turning the original model into a Quantized Neural Network (QNN). 
%
%Fixed point quantization helps to avoid the costly floating-point computation and reduces the memory footprint for storing both weights and activations, which is critical for resource-contrained platforms. Moreover, not all the MCUs feature a dedicated Floating point unit.
%
One of the most effective approaches \cite{hubara2017quantized} to quantize a real-valued weight parameter $w$ to a $Q$-bit signed fixed-point number $q(w)$ is by using the following quantization function:
\begin{equation}
\label{eq:quantization}
q (w) = clip_{[-1,1)}(2^{-( Q-1)} \cdot round(w \cdot 2^{(Q-1)}  )  \;,
\end{equation}
\noindent where $clip_{[a,b)}(x) = max(a,min(x,b))$.
We define then the integer $W = q(w) \cdot 2^{(Q-1)}$ as the corresponding INT-Q representation of $w$. According to \cite{hubara2017quantized}, the quantization rule (\ref{eq:quantization}) applies also to any activation value.
In this work, we explore the case of INT-8, INT-4, INT-2 and INT-1 data types as they are the most natural ones to fit in a 32-bit register of the targeted MCUs. 
If both weights and activations are INT-Q values, the convolution becomes a sum of products operation in the integer domain:
\begin{equation}
    \phi(w,x) = 2^{-2(Q-1)}\sum_{i \in C}W_iX_i \doteq    2^{-2(Q-1)} \cdot \Phi(W,X)\;. 
\end{equation}
\noindent where $C$ is the number of input channels and $\phi$ is the convolution operation.
$\Phi(W, X)$ is the accumulator value with high precision, i.e. INT-32 for INT-8 operands and INT-16 for sub-byte (INT-4, INT-2, INT-1) operands.
%
%As the target architecture provides for SIMD instructions up to support 8bit integer operands, to be able to exploit those also for INT-4 and INT-2 operands, a procedure of unpacking the input feature map pixels and filter weights and cast them to INT-8 is needed, as well as a function to pack back the results to the correct INT-Q representation.
%
To produce an output activation value, the accumualtion is compressed back into Q bits, working as input for the next layer.
% 8 BIT
For INT-8 data we adopt the compression approach proposed by Lai~et~al.~\cite{lai2018cmsis}, which relies on scaling and clamp operations, while for the 2 and 4 bit cases a thresholding-based 
\footnote{
The $\tau_p$ thresholds absorb bias, batch normalization and the $2^{-2(Q-1)}$ factor. Specifically, considering the batch-normalized $y= \gamma / \sigma (b + \phi - \mu) + \beta$ (where $b$ is the bias, $\gamma,\,\sigma,\,\beta,\,\phi $ are the batch normalization parameters), the thresholds are \begin{equation}
    \tau_p = [2^{Q-1} (p \cdot \sigma/\gamma- 2^{Q-1} \cdot (b - \mu) + \beta \cdot \sigma/\gamma]\;.
\end{equation}}
compression is considered, described by the staircase function that generalizes (\ref{eq:quantization}):
\begin{equation}
    Y = q\big(\phi(x)\big) = \sum_{p=-2^{Q-1}}^{2^{Q-1}-1} \left( p \cdot \chi_{[\tau_p, t_{p+1})} \cdot \Phi(W,X)\right)\;,
\end{equation}
\noindent where $\chi_s(\cdot)$ it the characteristic function of the interval $s$.
In this equation also the threshold values feature high precision (INT-16), since they are meant to be compared with INT-16 accumulations.
The staircase function is optimally implemented through a balanced binary tree where an INT-16 comparison takes place at every node.
%
%As the previous formula suggests we have $2^{Q}-1$ thresholds for each channel of the output feature map. 
To produce a Q-bit output, $2^{Q}-1$ threshold values per channel must be stored for any convolution layer. 
% 1 BIT
%The INT-1 format is a special case. The binarization can be redefined as $W=sign(w)$.
%
%By representing the $-1$ value with bit zero and the value $+1$ with bit one, the convolution reduces to a binary XNOR followed by a popcount operation:
The INT-1 format, where activation and weight values are expressed by binary values, is a special case because the convolution can be reduced to a logical XNOR and a bit-count operation:
\begin{equation}
    \Phi_{bin}(X) = \mathrm{popcount}( W\,\mathrm{xnor} X)
\end{equation}
where ${popcount}(\cdot)$ is the bitcount operator.
Also in this scenario, a thresholding procedure is applied for compression. %Threshold values (one INT-16 parameters per output channel) still accounts for bias and batch normalization.
%
%For INT-1 data we need only one INT-16 threshold, as the output will be in binary form.
%
% The threshold still requires higher precision (16 bits).
%

On the model accuracy side, it has been demonstrated that, through specific re-training techniques, the accuracy drop-off of quantized fixed-point networks can be significantly reduced \cite{hubara2017quantized, jacob2018quantization, rusci2019memory}.
Choi et al. \cite{choi2018pact}, for example, have proved that a 4-bit quantization leads to an accuracy level close to single-precision floating point representation.
The accuracy drop is limited to $3 \%$ when running ResNet50 on Imagenet with 2-bit weights and 4-bit activations and to $6.5 \%$ when downscaling the weights and activations to 2 bits.
Furthermore, the authors of \cite{moons2017minimum} investigated the trade-off between energy efficiency and accuracy of QNNs, highlighting the practical effectiveness of the sub-byte fixed-point networks. % in terms of reduction of the model size and in terms of computational energy efficiency boost in the inference.
At the cost of specific retraining procedures, the accuracy drop of is kept very close to the single-precision floating point counterpart while the energy efficiency gain, at the iso-accuracy, is orders of magnitude higher.
Moreover, for the investigated networks, trained on CIFAR-10 and MNIST datasets, the energy consumption achieved with 1- to 4-bit fixed-point networks, at iso-accuracy, outperforms the 8-bit counterpart by up to 10$\times$.  

%%CMSIS NN DESCRIPTION %%%
\subsection{Dataflow Schedule and Data Layout}
\label{sec:dataflow_layout}
\begin{figure}[t]
  \centering
  \includegraphics[width=\linewidth]{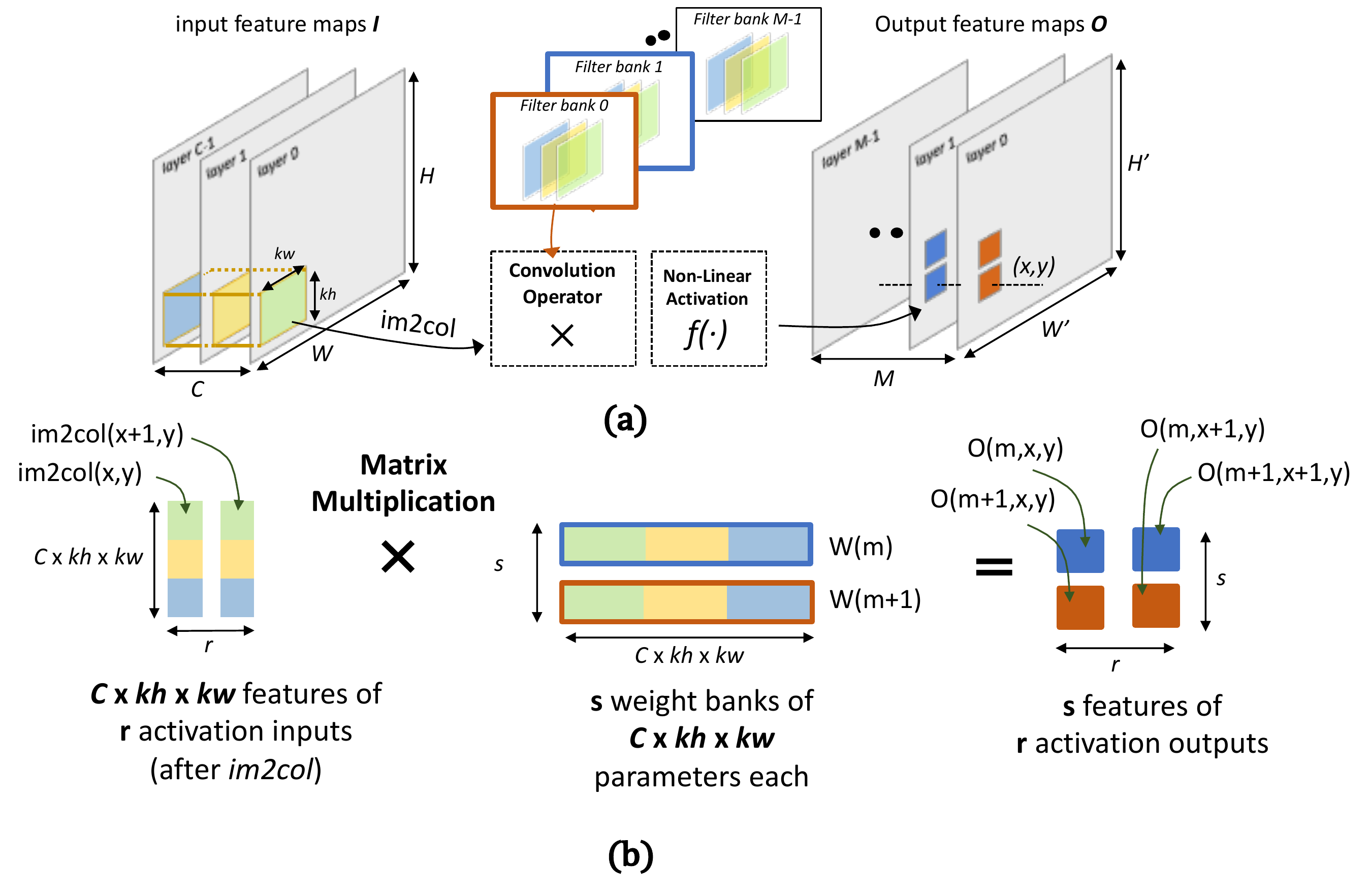}
  \caption{(a) Dataflow of the spatial convolution kernel (b) Convolution inner loop computation as a matrix multiplication.}
  \label{fig:cmsis}
\end{figure}

In this subsection, we detail the dataflow schedule and data layout as implemented in the CMSIS-NN library \cite{lai2018cmsis}, which is at the base of the proposed library.
%
%The convolution layer extracts a new feature map by computing a dot product between the filter weights and a small receptive field in the input feature map.
A convolution layer, standing as the basic building block for a CNN or a QNN model, produces an output feature map based on a set of weight filters and the output from the previous layer.
An activation value of any output feature map is computed as the dot product between a weights filter bank and a region of the input feature map, i.e. the $C$ features values of every point under the area $kw \mathrm{\ x\ } kh$ of the filter. To efficiently implement this operation on an MCU-like device, the convolution is decomposed into two phases: 
an \textit{im2col} step to load the input features of the current convolution into a contiguous memory array and a dot product. Besides the memory requirements of the activation maps and the model parameters, the \textit{im2col} demands an extra memory footprint of $ C \mathrm{\ x\ } kw \mathrm{\ x\ } kh$ values, on which the dot product operates.
%the first is oriented to the construction of a \textit{im2col} buffer which consists of transforming the image-like input into columns, and the second is devoted to the implementation of a matrix multiplication kernel.
Fig \ref{fig:cmsis}(a) shows graphically this operation.
Given this, the computation of one value of the output feature map, indicated as $O(m,x,y)$ becomes:
%\begin{equation}
%\label{eq:output_pixel}
%    O(m, x, y) = \sum_{k \in C} \sum_{i,j \in K} W_{i,j,k}(m) \cdot im2col_{i,j,k}(x,y)\;,
%\end{equation}
\begin{equation}
\label{eq:output_pixel}
    O(m, x, y) = \mathrm{dot}\Big( W(m) , \; im2col(x,y) \Big)\;,
\end{equation}
\noindent where $W(m)$ is the $m$-th bank of weight filter, im2col is the unrolled input buffer of length $C \mathrm{\ x\ } kw \mathrm{\ x\ } kh$.
%
% troppi dettagli...
%In addition, $0<m<M$, $0<x,y<E$, being M the number of channels in the output feature map and E its spatial dimension.
%
The inner loop of the convolution dot product is realized through a matrix multiplication kernel, as depicted in Figure \ref{fig:cmsis}(b). In general, $s$ output features of $r$ activation outputs ($s$=2 and $r$=2 in the example in figure) can be computed at this low-level stage. As a specific case, CMSIS-NN implements a matrix multiplication kernel working on two spatially adjacent pixels of two consecutive channels inside the inner loop of the convolution kernel; we identify this configuration as 2$\times$2, as explained in detail in Section \ref{sec:size_explo}.

%\noindent where $O(c,x,y)$ is the $(x,y)$ pixel of the $c$-th channel of the Output feature map, $im2col(x,y) = X(Sx-i, Sy-j, k) $ and $W(c) = W(c,i,j,k)$ are the im2col buffer and the subset of weights necessary to compute the specific output pixel, $w(c,i,j,k)$ is a filter weight at specific location and $X(Sx-i, Sy-j, k)$ is a pixel of the input feature map. 
%
%In addition, $0<c<M$, $0<x,y<E$, $M$ and $E$ are the numbers of channels and the spatial dimension of the output feature map, $S$ is the given stride size, whereas $K$ and $C$ are the spatial dimension of the kernel (i.e., the spatial receptive field of the filter) and the number of input channels, respectively.
%
Moreover, authors of \cite{lai2018cmsis} demonstrated the most convenient data layout to be  Height-Width-Channel (HWC), as it introduces minor overhead when building the im2col buffer with respect to the Channel-Height-Width (CHW) layout. %according to which
According to such a layout, the data along the channels is stored with a stride of 1, data along the width is stored with a stride equal to the number of channels $C$.

\subsection{Target Architectures}
\label{sec:target_architecture}

The target architecture of this work is based on a Parallel Ultra-Low-Power (PULP) cluster of RISC-V based processors.
A commercial embodiment of this architectural template is GAP8~\cite{flamand2018gap}, on which we run our experiments.
The GAP8 PULP cluster contains eight RISC-V cores, implementing a 4 stage in-order single-issue pipeline, supporting the RV32IMC instruction set \cite{waterman2016risc}, plus extensions targeting energy-efficient digital signal processing and machine learning (Xpulp) \cite{gautschi2017near}.
The cores are served by a 64kB L1 data memory, named Tightly-Coupled Data Memory (TCDM), enabling shared-memory parallel programming models such as OpenMP.
The shared L1 can serve all memory requests accessing different banks in parallel with single cycle access latency.
The 4 KB cluster program cache is also shared among the cores \cite{loi2018quest}.
The cluster is also provided by an Event Unit which manages synchronization and thread dispatching, enabling low-overhead and fine-grained parallelism, thus high energy efficiency:
%The cluster is provided also by an HW block, the Event Unit, which manages synchronization and parallel thread dispatching, enabling low-overhead fine-grained parallelism and high energy efficiency in parallel workloads.
%
each core waiting for a barrier is brought into a fully clock gated state.
%, zeroing its dynamic consumption.
%to maximize efficiency in fetching data-parallel code and implemented targeting energy efficiency \cite{loi2018quest}.
%
The cluster features also a DMA controller which manages the transfer between the L1 and the L2 memory (512kB in size), the latter residing outside-of-the-cluster of the GAP-8 architecture.
The Xpulp extensions available in the ISA\footnote{https://github.com/pulp-platform/riscv/tree/master/doc} include hardware loops, load/store with post-increment, Multiply and Accumulate as well as dedicated digital signal processing extensions inferred in the c code as built-in functions, presented below.
%The ISA extensions include hardware loops to accelerate \textit{for} statements, load/store with post-increment, Multiply and Accumulate as well as Single Instruction Multiple Data (SIMD) instructions, bit manipulation instructions and lightweight support for fixed-point operations such as saturation and clipping. The ones of particular interest in this work are presented below.

The SIMD vectorial instructions allow processing more sub-word data in parallel, most of them taking only one clock cycle.
The vectorial data types to be used with such instructions are \textit{v4s} and \textit{v2s}: 
\textbf{\textit{v4s}} allows to fill a 32bit register with four INT-8 data, \textbf{\textit{v2s}} does the same by filling the register with two INT-16 integers, in one clock cycle.
%
%For example by casting \textit{an} INT-8 pointer to \textit{v4s}, we can put in a 32bit register four consecutive INT-8 operands, which are ready to be processed in SIMD-like style.
%
%This operation takes a single load instruction, allowing to save many loads when processing sub-word data.
%
Sum of dot products SIMD instructions are provided to process either two 16 bit (\textit{sdotp2}) or four 8 bit  (\textit{sdotp4}) integer operands in a single cycle.
\textbf{\textit{sdotp4}} takes two \textit{v4s} data operands as input and computes the sum of dot products over the same accumulator, which is the INT-32 output of the built-in function.
The \textbf{\textit{max4}} instruction instead allows to compare two \textit{v4s} operands by returning the element-wise maximum, in one cycle.
%
%More specifically the output, returned as \textit{v4s} type, contains four INT-8 elements, each corresponding to the maximum value of the same positioned byte of the two input operands.

%\subsubsection{\textbf{Bit manipulation instructions}}
\textbf{\textit{bextract}} extracts, in one clock cycle, a specified number of bits ("size") from a register, starting at a specified position ("offset").
%\textbf{\textit{bextract}} allows extracting a specified number of bits from a variable, with a single clock cycle of latency.
%
%Two parameters can be passed to the function as immediates: the "size" parameter specifies the number of bits to be extracted from the source, and the "offset" parameter specifies the starting point of the extraction.
%
The extracted bits are then sign-extended and stored in the destination register.
The natural counterpart is the \textbf{\textit{bitinsert}} built-in function, specifying the number of bits to be inserted ("size") to the destination register, starting from the specified position ("offset").
%As before, two parameters passed to the function as immediates allow to specify the number of bits (of the source) to be inserted ("size") (to the destination variable) and the starting position of the insertion ("offset").
%
%The input and the output of the functions are 32bit integers.
%
\textbf{\textit{pack4}} allows to pack four INT-8 variables in a SIMD \textit{v4s} data type in two clock cycles.
%
%It takes four variables as input and returns the packed \textit{v4s} as output.
%
Finally, the \textbf{\textit{popcnt}} built-in returns, in one cycle, the number of bits set to one in a word which is passed to the function as input.
%Finally, the \textbf{\textit{popcnt}} built-in instruction is equivalent to the pure GCC sequence \textit{popcount}. 
%
%In one cycle, it returns the number of bits set to one in a word which is passed to the function as input.
%% PULP NN LIBRARY

%In this section we present an open-source optimized library which contains a full set of kernels and utilities to efficiently run Quantized Neural Networks (QNNs) on a RISC-V ultra-low-power multicore cluster, targeting all the INT-Q data representations presented above. 
%

%First we show the RISC-V implementation of the library and its optimization by fully exploiting the ISA at our disposal, which leads to speed up the computation and energy saving; as second we present the parallel programming model used to enable the multicore execution of the kernels, which leads to a near linear speed-up with respect to the single core execution. In the end, we present a convolution kernel exploration which consists of a study of the kernel sizes to find the optimal data reuse condition, which directly affects the output pixels computation throughput.
%

\section{PULP-NN Library}
\label{sec:PULPNN}
This section introduces the PULP-NN library and describes the optimization of the kernels with the presented RV32IMCXpulp extended ISA on a parallel cluster of eight processors and the optimization of the main computational kernel of the library: the matrix multiplication. We focus on the computational part since we are interested in exploring software solutions capable of achieving high computing performance and energy efficiency, on top of parallel edge architectures like PULP.
%%% IMPLEMENTATION ON RISCV %%%
\subsection{Implementation and Optimization on RISC-V}
\label{sec:riscv_imple}

We present implementation details of the most significant QNN kernels on the target RV32IMCXpulp ISA. The experiments are conducted assuming that all the data resides in L1 memory of the PULP cluster.% Avoiding the insertion of an additional degree of freedom into our optimization procedure (which would consist of memory management) allows us to better understand the computational limits of the kernels and to investigate focused solutions to obtain a high performance and high energy efficiency software computing library, built on top of a parallel MCU architecture.
%

%%% CONV LAYER IMPLEMENTATION ON RISC-V%%%
\subsubsection*{\textbf{INT-8 Kernels}}
\begin{figure}[t]
  \centering
  \includegraphics[width=0.9\linewidth]{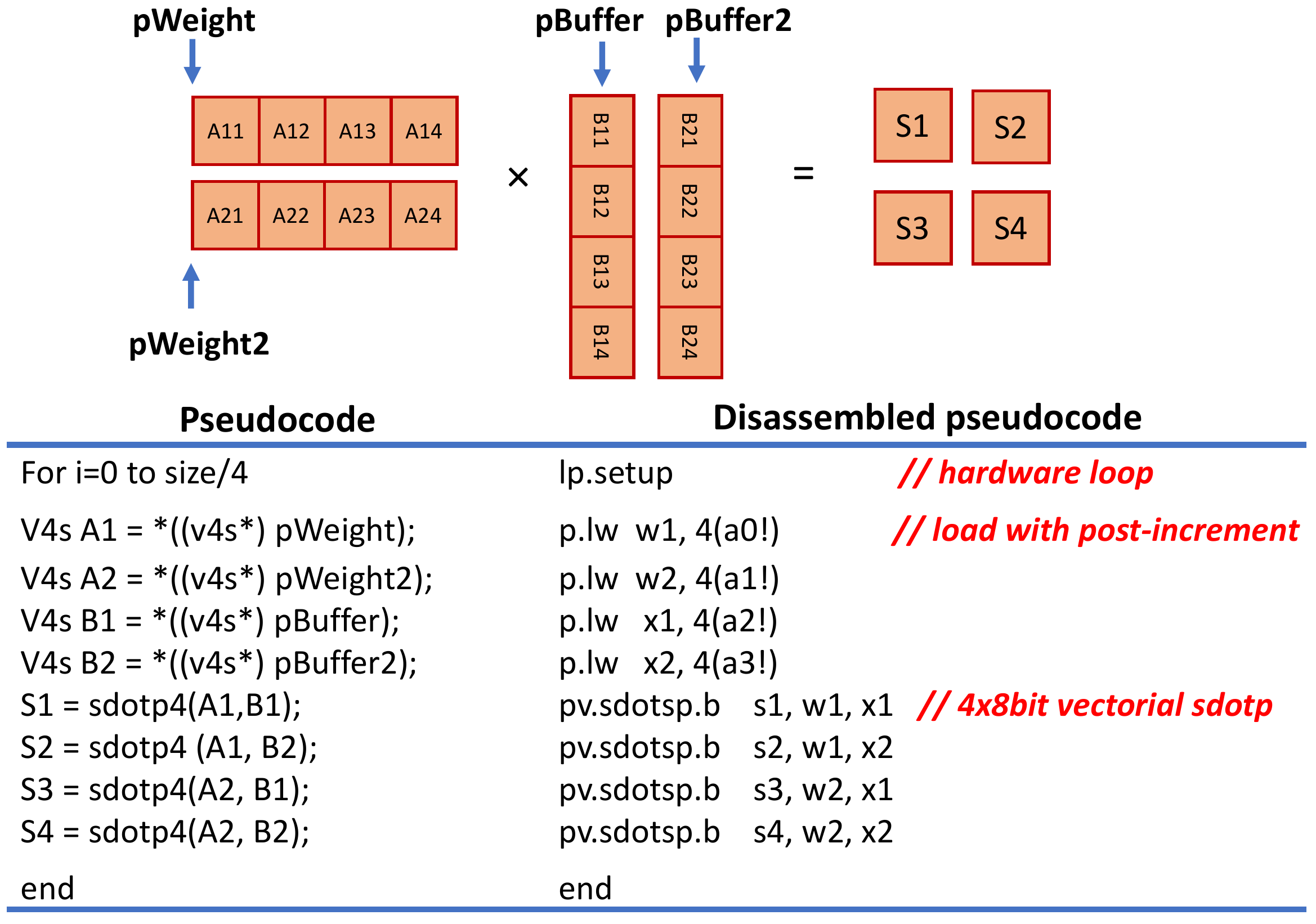}
  \caption{2$\times$2 sized matrix multiplication kernel for INT-8 data operands.}
  \label{fig:2x2kernel_isa}
\end{figure}

The first layer for which we detail the implementation is the convolution one.
We first consider the INT-8 kernel, as it also provides a basis for the implementation of INT-4 and INT-2.
%The target data type is INT-8, as it is well supported by the ISA extensions, and is also used for sub-byte data operands, except for INT-1.
%
%The same INT-8 core kernel is also used for sub-byte data operands, except for INT-1, as we want to take full advantage of the ISA.
%
%In such cases additional functions to unpack and cast and pack data are needed, and they will be presented subsequently in section \ref{sec:data_packing_unpacking}.
%
Starting from the implementation presented in section \ref{sec:dataflow_layout}, with a 2$\times$2 matrix multiplication kernel, we optimize it to fully exploit the RV32IMCXpulp ISA.
Since the matrix multiplication operation has to be looped over the size of each filter bank ($ C \mathrm{\ x\ } kw \mathrm{\ x\ } kh$), we take advantage of the \textit{hardware loops} to accelerate the \textit{for} statement.
In the inner loop, we also exploit the load and store with post-increment since the access pattern to the im2col and filter elements is extremely regular by construction.
In the same way, we use the 8-bit SIMD instructions to increase the throughput of the computation.
%
%Keeping in mind the data layout presented in section \ref{sec:dataflow_layout}, we now focus on the implementation of the  $2x2$ convolution kernel for INT-8 data operands on the targeted RISC-V based architecture.
%
Figure \ref{fig:2x2kernel_isa} graphically schematizes the execution of the inner loop of the matrix multiplication kernel and reports the corresponding assembly code.

%We demonstrate that, by fully exploiting the targeted ISA, it can be possible to increase the throughput by almost $9x$ on a single core execution with respect to the plain implementation presented above.

After filling two im2col buffers that are needed to compute two spatially adjacent output pixels, the matrix multiplication inner loop takes place as follows.
At every iteration of the loop, four consecutive elements are loaded into the register file from each of the two im2col buffers (pointers \textit{pBuffer1} and \textit{pBuffer2} in the figure), and from two weight banks (pointers \textit{pWeight} and \textit{pWeight2}), after casting INT-8 pointers to \textit{v4s}.
The total number of load operations required is four.
In this way we have sufficient elements to set four \textit{sdotp4} built-in functions over four different accumulators.
%, which will correspond to the output pixels.
%
%When using the SIMD extensions, we will talk about the sum of dot products to loads ratio instead of MACs to loads, since this measure is ISA independent.
%
Hence, in a single run of the inner loop of the matrix multiplication kernel, we can compute four sdotp4 instructions, which correspond to 16 MAC operations, at the cost of four load instructions. 
%
%Considering a single run of the inner loop, we are now able to perform four vector sdotp4 instructions which translate in 16 MACs at the cost of four loads.
%
%The possible leftover due to some mismatch between the convolution layer size and matrix multiplication kernel size is handled with a plain implementation of the kernel.
%
%The experimental results and the comparison with the plain implementation are reported in section \ref{sec:RISC-V_impl_res}.
%

%%%FULLY CONNECTED KERNEL %%%
Since the fully connected kernel is a simple matrix by vector multiplication, the previous methodology naturally scales to it.
Here there is no need to build the im2col buffer since the spatial dimension of the filters is the same size as the spatial dimension of the input feature map.
%
%The output feature map produced by a fully connected kernel has neither height or width dimension, which we can consider equal to $1$. 
%
%For this reason, the previous reasoning about data reuse to compute more output pixels along the spatial dimension is no longer valid.
%
%What we can still do is 
To reduce load instructions and exploit a data reuse mechanism, the fully connected kernel implements $2x1$ matrix multiplication kernel within the inner loop (see Section \ref{sec:size_explo} and Figure \ref{fig:kernel_size}).
%To exploit data reuse along channels, by implementing $2x1$ matrix multiplication kernel (see Section~\ref{sec:size_explo}).
%
By loading two different subsets of weights, we can compute two consecutive output pixels along the channel dimension.
By using the SIMD ISA extensions as before, with three loads we are able to set two \textit{sdotp4} vector operations per loop cycle, which translates in 8 MACs.

Ancillary operations also take benefit of the DSP extensions.
ReLU, which consists of a simple $\max$ looped over the input feature map, exploits \textit{hardware loops}, load store with post-increment and the SIMD \textit{max4} built-in instruction.
%to compare a vector mask of zeros.
%
The same is also used to optimize the max-pooling kernel, which is implemented in two steps: first along the width dimension, working destructively \textit{in situ} on the input buffer; then along the height dimension.
%
%%SUPPORT FUNCTIONS%%
\subsubsection*{\textbf{Sub-byte Extensions}}
\label{sec:data_packing_unpacking}
\begin{figure}[t]
  \centering
  \includegraphics[width=0.9\linewidth]{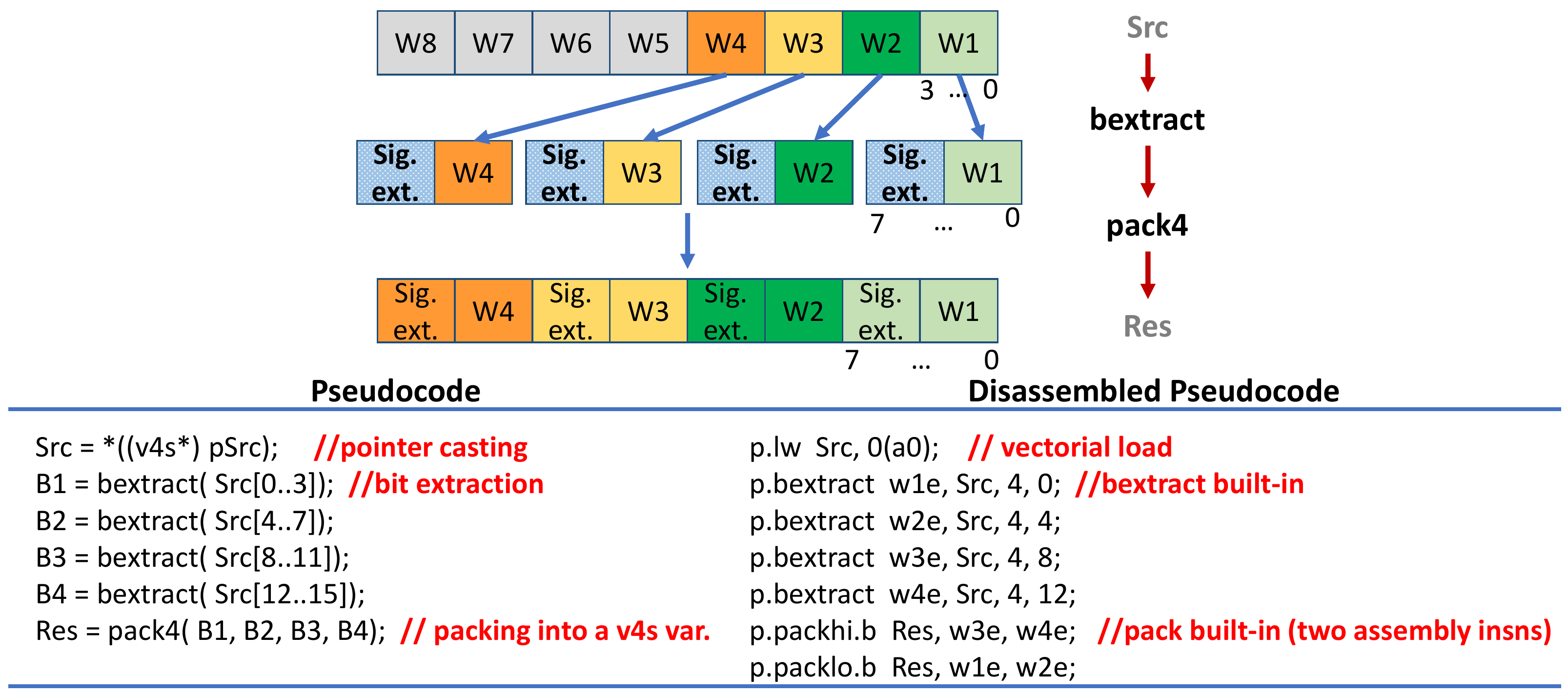}
  \caption{INT-4 to INT-8 unpacking function.}
  \label{fig:int4toint8}
\end{figure}

The smallest data type well supported by the ISA with the SIMD extensions is INT-8.
To exploit efficiently such vector operations, it is necessary to provide additional support functions to convert sub-byte data, i.e. INT-2 and INT-4, into INT-8.
Having sub-byte operands compactly stored in memory, in the case of INT-4 data two consecutive elements are placed in a single byte.
%We assume that the sub-byte operands are stored compactly in memory, e.g., two consecutive INT-4 elements are stored in a single byte.
%
%For example, considering INT-4 data representation, we expect that two consecutive pixels or weights are stored in a single INT-8 variable.
%
The casting operation, realized through the  \textit{pulp\_nn\_int4\_to\_int8} function, takes place either when building the im2col buffer as well as in the innermost loop of the matrix multiplication kernel to "unpack" weight elements. To reduce the overhead due to the unpacking operations, combined use of the \textit{bextract} and \textit{pack4} built-in functions allows to extract four INT-4 elements (weights or pixels) with few instructions, as shown in Figure \ref{fig:int4toint8}. 
%
%First, we cast the pointer of the compressed INT-4 weights/pixels to $\textit{v4s}$, to have all the data to be extracted in a single register, reducing so the number of loads to be performed. 
%
After loading eight INT-4 data with a single load, four elements are extracted by means of the \textit{bitextract} built-in and packed into one single SIMD \textit{v4s} variable, which feeds the matrix multiplication kernel.

\begin{figure}[t]
    \centering
    \includegraphics[width=0.9 \linewidth]{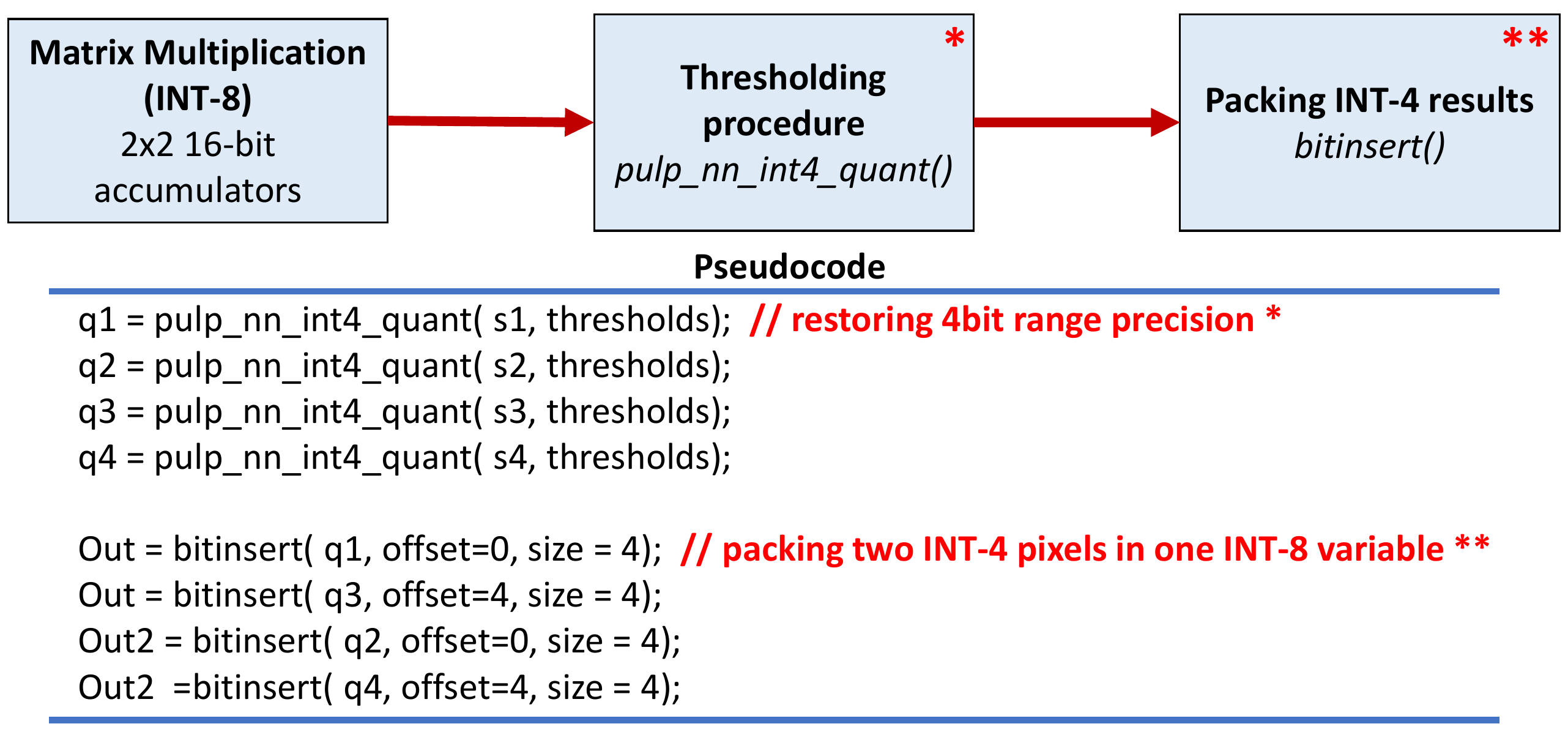}
    \caption{The compression procedure for INT-4 data types.}
    \label{fig:int4compression}
\end{figure}

The results of the matrix multiplication kernel (which is always performed with the INT-8 data type) are 16-bit long, as the accumulator features a precision higher than operands, as described in Section \ref{sec:background_quant}.
A compression procedure is thus needed to bring the result back to INT-4.
%
%To do this, we implement \textit{pulp\_nn\_int4\_quant} function which compares the 16 bit result with $2^4-1$ thresholds.
% 
Starting from the considerations in \cite{rusci2018work}, the 16-bit accumulator is compared with the corresponding $2^4-1$ threshold values, using an optimal balanced binary tree function, named \textit{pulp\_nn\_int4\_quant}.
Such a procedure is necessary to restore the precision of the results in a 4 bit range.
%cd 
To save memory footprint, two consecutive output INT-4 data are stored in a single-byte variable using the \textit{bitinsert} built-in function.
A graphical explanation of the compression mechanism is provided in Figure \ref{fig:int4compression}.
A similar process is implemented for INT-2 convolutions, by featuring dedicated \textit{packing} and \textit{unpacking} functions.
%The \textit{packing} and \textit{unpacking} process for INT-2 operands follows the same flow as INT-4. 
%
%In this situation it is not needed to cast the INT-2 pointer to \textit{v4s}, since four INT-2 operands fit also in INT-8 variable.
%
%The compression is performed in the same way as INT-4.
%
%%% 1 BIT %%
\subsubsection*{\textbf{Binary Convolution Kernel}}
For the INT-1 data representation no casting/unpacking is needed because of the natural support provided by the ISA for binary operations.
We exploit the bitwise instructions to implement the convolution kernel, which is based on bitwise XNOR operations between binary weights and binary inputs.
The accumulator is filled by counting the number of ones occurring after the XNOR.
To this purpose we use \textit{popcnt} built-in.
The 16-bit accumulator is compared with a single threshold and results either in a zero or one, stored back into memory by means of the \textit{bitinsert} built-in function.

%%% MULTICORE EXECUTION %%%
\subsection{Multicore Execution}

\begin{figure}[t]
    %\centering
    \includegraphics[width=0.95\linewidth]{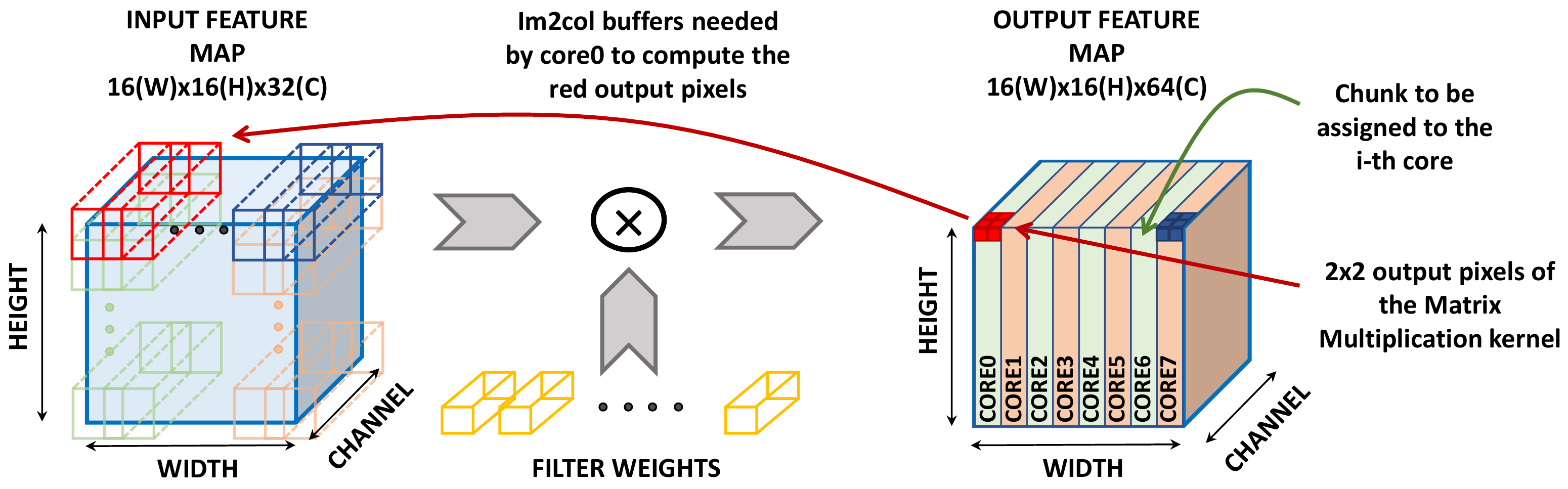}
    \caption{The right side of the figure shows how the chunks are assigned to the 8 cores of the PULP cluster. To take advantage of the HWC data-layout each chunk is built along the spatial dimension of the output feature map. The left side gives a graphical intuition of the need each core has to create its private im2col buffer. Considering the 2$\times$2 matrix multiplication kernel each core requires two private buffers of such type. }
    \label{fig:convdataparallel}
\end{figure}
% In this section we present a multicore execution of the kernels, providing for some implementation details.
%
%The convolution kernel and the fully connected kernel multicore execution implementation follows the same workload split among cores.
%
%We deepen the implementation of the convolution kernel, leaving aside the other one to avoid overloading the discussion.
%

%% CONV KERNEL PARALLELIZATION%%
As discussed above the \textit{convolution kernel} execution consists of two phases: the im2col function and the matrix multiplication kernel.
The proposed data-parallel multi-core optimization is motivated %takes advantage 
by the HWC format used to store pixels and weights and by the two phases of the dataflow.
Because of the HWC format, it is convenient to split the workload along the spatial dimension of the output feature map, in a way that each core computes the full set of $M$ output features for a given output spatial coordinate, as shown in Figure \ref{fig:convdataparallel}.
%a spatial portion of pixels, for all the channels of the output feature map.
%
%\cc{The computation of an output pixel requires to store the associated im2col buffer to be built from input feature map and then multiplied with a specific subset of weights, depending on the channel considered.}
%
To implement this strategy, each core requires a private im2col buffer.
More specifically, if we consider the 2$\times$2 kernel, each core must allocate and load two im2col buffers before running the matrix multiplication kernel.
Therefore, the parallelization boost comes at the cost of a small amount of additional memory footprint for the extra im2col buffers, which in the worst case (eight cores configuration) is about $9\%$ of the total when considering 16$\times$16$\times$32 sized input feature map, 16$\times$16$\times$64 sized output feature map and 64$\times$3$\times$3$\times$32 sized 3D convolution filter.
%
%On the contrary, this solution avoids the TCDM contentions we would have in the matrix multiplication kernel when fetching input pixels.
%
%Each core will access different memory locations to fetch pixels from the private im2col buffers.
%
% FIXMEEEEEEEEEEEEE 
%%To avoid unnecessary data replication, the pixels in the buffers do not overlap.
%
The weights instead are shared among the cores.
%
%In the matrix multiplication kernel, the possible TCDM contentions are managed in hardware.
%
%The TCDM logarithmic interconnect try to alleviate the ovearhead by disaligning the memory accesses when a TCDM contention occurs.
% 
%\begin{figure}[h]
%  \centering
%  \includegraphics[width=0.6 \linewidth]{multicore_chunck_4cores.jpg}
%  \caption{Example of how workload is split among cores. As an example here a 4 cores configuration is considered. The chunks are built along spatial dimensions on the output feature map.}
%  \label{fig:multicore_chunck}
%\end{figure}
%
% The solution proposed is efficient and allows us to achieve a near linear speed-up with respect to a single core execution.
%

Since the fully connected layer generates a set of neurons as output (i.e., the output feature map does not extend along any spatial dimension), the only dimension along which we can split the workload is the channel.
We assign a balanced number of neurons to be computed to each core.
%
% However, the number of output neurons usually is not so large; achieving high speedup is not easy because it is difficult to balance the workload on eight cores.
%
%This means that all the filter weights and the input pixels are shared among the cores and the possible TCDM contentions are managed in hardware.
%
The parallelization of the ReLu and the Max Pooling kernel is straight-forward: the chunk to be assigned to each core is a balanced group of pixels along the entire input feature map.
%
%
%The function compare a mask of zeros with the input pixels and it is to be looped over the input feature map.

%%MAX POOLING %%
%The Max pooling kernel requires more attention to the synchronization among the cores to keep the data and memory consistency.
%
%As defined before, the max pooling implementation is split into two phases.
%
%The first, which is made in \textit{situ}, performs the max pooling along x-axis. 
%
%On the result we perform max pooling along y-axis.
% 
%In a multicore execution environment we must take care of the synchronization among cores.
%
%In the first step we can set the multicore execution by splitting the workload along one spatial dimension.
%
%Only after each core has finished this step we can set the multicore execution along y-axis. 
%
%This would require an additional synchronization barrier which, in the end, does not affect the performance.
\begin{figure}[t]
  \centering
  \includegraphics[width=0.95\linewidth]{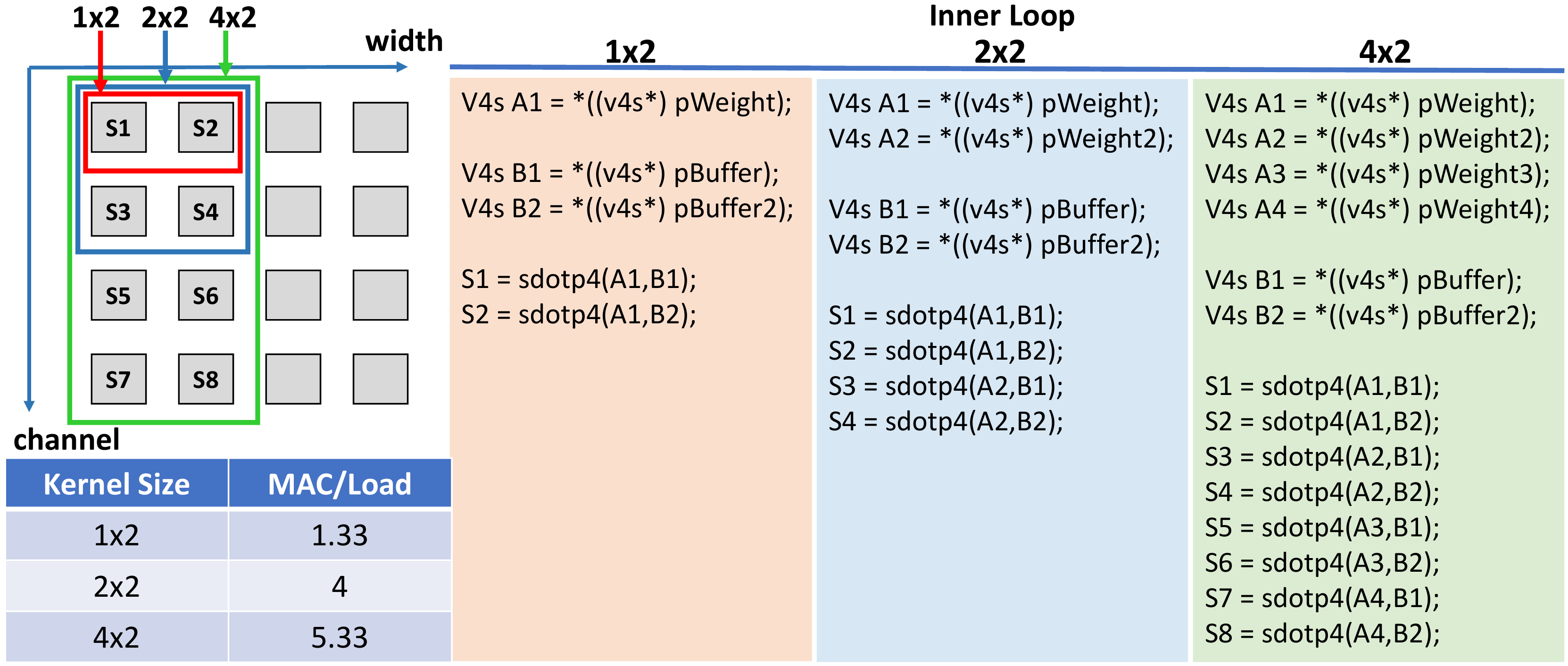}
  \caption{Inner loop of the matrix multiplication considering different sizes of the kernel.}
  \label{fig:kernel_size}
\end{figure}

%%%KERNEL EXPLORATION%%%
\subsection{Matrix Multiplication Kernel Size Exploration}
\label{sec:size_explo}
%
%To further improve the performance of the convolution layer, expressed as MACs/cycle, we optimized the matrix multiplication kernel which consists of a loop of MAC operations.
% by exploiting the data reuse at the register file level.
%
%Such a kernel consists of a loop of MAC operations.
%
%The cost to set them is due to the number of loads necessary to bring the operands in the register file.
%The matrix multiplication kernel is made of a loop of MAC operations.
%
%The cost to set the MAC operations in the inner loop of the kernel is due to the number of loads necessary to bring the operands in the register file.
%
%For example, if in a single run of the inner loop we perform a single MAC (implementing thus 1x1 sized kernel) we must pay for two loads.
To further increase the throughput of a memory intensive kernel such as matrix multiplication, it is important to reduce the cost of loading the operands into the registers as much as possible, by maximizing the \textit{data reuse} at the register file level.
%To increase the throughput, it is important to reduce such a cost as much as possible, by maximizing the number of MAC operations we can set in the inner loop for a fixed number of loads, i.e., by maximizing \textit{data reuse} in the register file.
% the MAC to load ratio.
%

The direct implementation of the Equation~(\ref{eq:output_pixel}) would be inefficient since, from a computation perspective, %For example, the equation (\ref{eq:quantization}), from a computing perspective, is equivalent to set one MAC operation performed over the same accumulator, in the inner loop.
two loads are required (one to fetch an im2col element and one to fetch a weight parameter) to feed the MAC instruction.
%if in a single run of the inner loop we perform a single MAC (implementing thus 1x1 sized kernel) we must pay for two loads.
%
%This is inefficient since 
In this scenario, one load stall will be necessarily paid, degrading the IPC metric and reducing the throughput.
To avoid the stall, multiple output data can be computed within the inner loop of the dot product routine, i.e., the inner loop of the matrix multiplication kernel.
%, setting more MAC operations over different accumulators.
%
%Starting from equation (\ref{eq:quantization}) in section \ref{sec:dataflow_layout}, 

%If we want to compute an adjacent output pixel in position $(x+1,y)$, the equation (\ref{eq:output_pixel}) becomes:
When applying equation (\ref{eq:output_pixel}) to compute the output data at the spatial coordinate $(x+1,y)$, the formula becomes:
%\begin{equation}
%    O(m, x+1, y) = \sum_{k \in C} \sum_{i,j \in K} W_{i,j,k}(m) %\cdot im2col_{i,j,k}(x+1,y)\;,
%\end{equation}
\begin{equation}
    O(m, x+1, y) = \mathrm{dot}\Big( W(m) , \; im2col(x+1,y) \Big)\;.
\end{equation}
\noindent We can notice that the same subset of weights is used in the computation of the output data at coordinates $(x,y)$ and $(x+1,y)$ .
What changes is only the im2col buffer.
%
%The latter equation is equivalent to 
When operating on these two point simultaneously, the inner loop consists of two dot product operations, which are performed over two different accumulators. By reusing the register that stores the elements of $W(m)$ along the spatial dimension we can set two \textit{sdotp4} operations at the cost of one additional load (three in total), needed to fetch the elements of the second im2col buffer.
%The same reasoning applies when considering an inner loop of two MAC operations, performed over two different accumulators, corresponding to the two output data.
%
%In this case, we can reuse the register which stores the elements of $W(m)$ along the spatial dimension and set two \textit{sdotp4} operations at the cost of one additional load (three in total), needed to fetch the elements of the second im2col buffer.
%
So doing, we build the 1x2 sized kernel and increment the MAC to load ratio.
If extending this strategy also to the feature dimension, 
the inner loop of the convolution can operate on a 2$\times$2 sized kernel, i.e. computing four accumulations related to two features of two separate output pixels $(x,y)$ and $(x+1,y)$. Such a kernel size is the one used by ARM CMSIS-NN.
In this case, an additional subset of weights, $W(m+1)$ is needed and, at the cost of four loads, we can perform four \textit{sdotp4} operations in the inner loop.
By means of this upgrading, the MAC to Load ratio grows up to $4$.
% with respect to the $xxxx$ case.
%
% insert here the drawback
%

%The upper limit to the level of data reuse in the register file is due to the available registers to store operands and accumulators.
%The exploitation of the data reuse at register file level is limited by the resources available (i.e. the number of registers) to store the operands and the accumulators.
%
%We can further extend the procedure to build differently sized kernels to provide for a full exploration of the kernel size design space with the goal to find the best register file data reuse condition which maximizes the throughput.
%

Let us consider the 4x2 sized kernel, which means we want to compute two adjacent spatial pixels along four consecutive channels of the output feature map.
Following what we said before, we need to build two im2col buffers, and we need four different subsets of weights. 
The elements loaded in the register file are reused similarly as presented before to maximize the MAC to Load ratio.
Figure \ref{fig:kernel_size} explains the concept of register file data reuse.
%
%Following what we said before, we need to build two im2col buffers, which will be reused along channel dimension and we need to use four different subsets of weights which will be reused along the spatial dimension.
%
As a counterpart, we can explore the 2x4 sized kernel.
In this case, the reasoning is reversed.
%: two different subsets of weights are needed and reused along the spatial dimension, and four im2col buffers have to be build and reused along channels.
%
The MAC to load ratio we can achieve in both cases is $5.33$, as we compute 32 MACs at the cost of 6 load operations, in a single run of the inner loop.
Thus we expect a better throughput with respect to the 2$\times$2 sized area.
It is interesting to notice that in the 2x4 case, the memory footprint is slightly higher than the 4x2 sized kernel because of the two additional im2col buffers.
%
%This could raise several memory problems when exploiting the parallelization over the eight cores of the cluster, as in this case, the footprint is eight times higher.
%
For the same performance, the former is thus to be preferred between the two.
%
% We can continue in this exploration by building the 4x4 sized kernel. 
% %
% With this structure we set the computation of 16 output pixels in parallel.
% %
% The data required to do this consists of four im2col buffers and four subsets of weights.
% %
% The ideal MAC to load ratio grows up to $8$, as with eight loads we can compute 64 MACs in a single run of the loop.
%

It is important to notice that the upscaling of the kernel size is limited by the resources available in the register file to store operands and accumulators, thus limiting the \textit{data reuse} design space at this level. We explore such a space to find the best register file data reuse condition which maximizes the throughput. The experimental results and further considerations are provided in Section \ref{sec:kernel_expl_results}.

%%% EXPERIMENTAL RESULTS %%%
\section{Experimental Results and Discussion}
\label{sec:results}
The solutions presented in this paper are evaluated on the off-the-shelf GAP8 \cite{flamand2018gap} microcontroller, which is an embodiment of the target PULP architecture with eight cores.
The same experiments can also be replicated on the open-source PULP platform\footnote{https://github.com/pulp-platform.} via RTL simulation.

\subsection{Comparison with RV32IMC ISA}
\label{sec:RISC-V_impl_res}
\begin{figure}[t]
  \centering
  \includegraphics[width=0.9\linewidth]{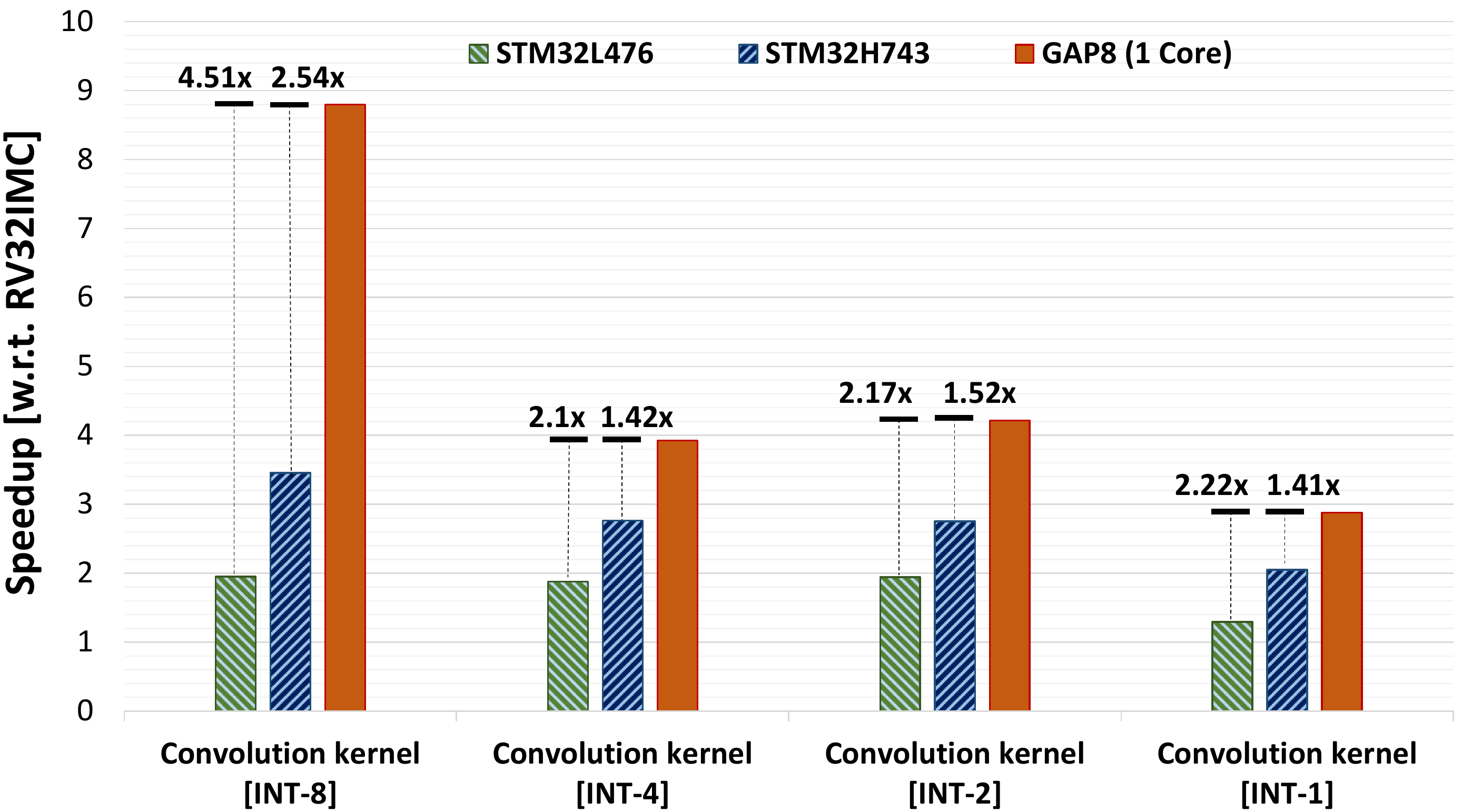}
  \caption{Speed-up of PULP-NN conv kernels (single core execution on GAP-8) and CMSIS-NN conv kernels (on STM32H7 and STM32L4) with respect to RV32IMC ISA.}
  \label{fig:plain_comparison}
\end{figure}

To evaluate the proposed library, which exploits the DSP extensions available on the RI5CY processor \cite{gautschi2017near}, we first compare the optimized single core execution of the convolution kernels with respect to a corresponding \textit{RV32IMC} ISA implementation, sweeping all the INT-Q datatypes supported.
This evaluation is performed by benchmarking a convolution kernel operating on a 16x16x32 input tensor (HWC data-layout) with a filter size of 64x3x3x32 ($C \mathrm{ x } kw \mathrm{ x } kh \mathrm{ x } M$).
We consider the convolution kernel as its workload is dominant when inferring an entire QNN (about 96 \% on the CIFAR-10).
As a second term of comparison, we run the kernels on off-the-shelf STM32H743 \cite{STM32H743xl}  and STM32L476 \cite{STM32L476} commercial microcontrollers based on ARM CORTEX-M7 and CORTEX-M4 cores respectively, using the CMSIS-NN \cite{lai2018cmsis} library.
To run the sub-byte quantized version of the convolution layer on such MCUs, we refer to \cite{rusci2018work}; the extension to the CMSIS-NN library is open access\footnote{https://github.com/EEESlab/CMSIS\_NN-INTQ}.
%
%a 64x3x3x32 convolution layer applied on a 16x16x32 input tensor and producing an output feature map of 16x16x64 pixels.
%
%
%The kernels considered in these experiments have the following parameters:
%\begin{itemize}
%    \item \textit{Convolution kernel}: it is made of a 64x3x3x32 sized 3D filter, applied to a 16x16x32 input tensor and generating an output feature map of 16x16x64 pixels;
%    \item \textit{Fully connected kernel}: the filter is 32x8x8x16 in size applied to a 8x8x32 feature map;
%    \item \textit{ReLu}: The activation function is performed over a 32x32x32 tensor;
%    \item \textit{Max Pooling}: taking a 32x32x16 feature map it performs the dimensionality reduction with stride $2$.
%\end{itemize}
%The convolution layer takes as input a 16x16x32 feature map and generates a 16x16x64 sized output feature map.
%
%The 3D filter is 64x3x3x32 in size; the amount of stride and padding is equal to $1$.
%
%The layer parameters of the remaining kernels are the following: the \textit{ReLu} activation is performed over a feature map of 32x32x32 pixels, wheres the pooling kernel is applied to a feature map of 32x32x16 pixels, with stride $2$.
%
%We profile the kernels by measuring the execution cycles.
%
The results of the comparison are presented in terms of speedup with respect to the \textit{RV32IMC} implementation and reported in Figure \ref{fig:plain_comparison}.

%Analyzing the results, we conclude that the ISA extensions increase the performance considerably.
%
%For INT-8 operands we achieve a speedup up to $4.51 \times$ with respect to the ARM-based processors since the latter do not have support for 8-bit operations.
%
%When running the optimized version of the convolution kernel we can observe a speedup of $8.8 \times$ with respect to the plain version and a speedup of $4.51 \times$ with respect to the Arm CMSIS-NN implementation of the kernel.
%
%Even better performance can be noticed when considering a fully connected layer, where we have a speedup that overcomes $10 \times$ with respect to the plain version.
%
%The execution speedup grows also for ReLu and Max Pooling functions where we use the built-in \textit{max4}.
%
%More specifically, in the first we observe a speedup of $11.9 \times$, whereas in the second, although with less impact, we notice a speedup of about $3 \times$.
%
We achieve the best speedup on the INT-8 convolution kernel, mainly thanks to the 8-bit SIMD \textit{sdotp} instructions. The ARM ISA features support for 16-bit instructions only, dividing by a factor of 2 the MAC throughput with respect to the RI5CY processor. Moreover additional rotate instructions are required on ARM architectures to pack 16-bit vector data to feed the MAC units \cite{rusci2018work}. Finally, hardware loops provide another factor of improvement with respect to ARM. Thanks to these extensions we outperform by $2.54 \times$ and $4.51 \times$ the STM32H7 and L4 MCUs respectively, despite the CORTEX-M7 processor available in the STM32H7 featuring a dual-issue pipeline.
%

%As regards the convolution kernel for INT-8 operands we achieve the best speedup thanks to the 8-bit SIMD instructions available on PULP architectures, which on the contrary are not supported by ARM CORTEX-M7/M4 ISAs.
%

%
When considering sub-byte data types, we notice a degradation of the speedup with respect to RV32IMC which passes from $8.8 \times$ (INT-8) to $3.69 \times$ and $4.22 \times$ for INT-4 and INT-2 data respectively.
%performance, as neither INT-4 nor INT-2 data operands are supported by the SIMD extensions.
%
%The speedup with respect to the \textit{RV32IMC} implementation passes from $8.8 \times$(INT-8) to $3.69 \times$ and $4.22 \times$ for INT-4 and INT-2 data respectively.
%
Such degradation is due to the additional instructions to unpack and cast INT-2/4 operands to INT-8 ones. Although these operations are implemented with \textit{bextract} and \textit{pack4} instructions, they do not achieve the same speedup as the INT-8 convolution kernel, limiting the overall speedup for sub-byte kernels, still leading to a speedup of $1.42 \times$ and $2.1 \times$ with respect to STM32H7 and STM32L4 for INT-4 kernel, respectively, and a speedup of $1.52 \times$ and $2.17 \times$ with respect to H7 and L4 for INT-2 kernel, respectively.
%
%The  bit manipulation instructions provided by the ISA allow us to achieve a 
%
The ARM CORTEX-M7/M4 processors do not have ISA support for efficient bit manipulation instructions nor for popcount instruction which is helpful for the INT-1 case. However most of the computational load of this kernel is implemented with xnor instructions available in all considered ISAs. Hence, the proposed implementation, runs $1.41 \times$ and $2.22 \times$ faster than the extended CMSIS-NN solution on STM32H7 and STM32L4 respectively.
%
%The gaps we have presented in this section may be reduced by the recently announced ARM-v8.1 extensions \cite{Armv8.1}, but they are not supported yet by any device.
%In the end, considering the binary convolution kernel, the \textit{popcnt} built-in allows us to achieve a speedup of $1.41 \times$ and $2.22 \times$ compared to the extended CMSIS-NN solution on STM32H7 and L4 respectively.
%
%The speedups with respect to the ARM-based processors are because the latter does not support 8-bit operations, do not have efficient bit manipulation instructions or support for the popcount instruction. 

%%%            MULTICORE EXECUTION RESULTS         %%%
%\begin{center}
    \begin{table*}[t]
    \centering
  \begin{tabular}{ccccccc}
    %\toprule
    Configuration& Nr. &  I\$ stall & TCDM cont.& Load stall& Total exec.  & Speedup\\
    %\toprule
      & insns &cycles&  cycles&cycles&cycles& \\
    \hline
    \hline
    \\
    \textbf{Convolution} &&&&&& \\
    1 CORE & $2546k$ &  $1.3k\,\, (0.05\%)$ & $0$ & $18k\,\,(0.7\%)$ & $2586k$ & $1 \times$\\
    2 CORES & $1286k$ &  $4.5k\,\, (0.35 \%)$ & $1.4k\,\, (0.11 \%)$ & $11k \,\, (0.85 \%)$ & $1299k$ & $\mathbf{1.99 \times}$\\
    4 CORES & $636k$ &  $5.7k\,\, (0.86 \%)$ & $3.8k\,\, (0.56 \%)$ & $5.5k \,\, (0.83 \%)$ & $660k$ & $\mathbf{3.92 \times}$\\
    8 CORES & $318k$ &  $21.5k\,\, (5.96 \%)$ & $6.6k\,\, (1.83 \%)$ & $2.7k \,\, (0.75 \%)$ & $361k$ & $\mathbf{7.16 \times}$
    \\
    \hline
    \\
    \textbf{Fully connected}&&&&&& \\
    1 CORE & $20.7k$ &   $0.03k \,\, (0.09 \%)$ & $0$ & $0$ & $33k$ & $1 \times$\\
    2 CORES & $10.4k$  & $1.1k \,\, (6.25 \%)$ & $1k\,\,(5.69\%)$ & $0$ & $17.6k$ & $\mathbf{1.89 \times}$\\
    4 CORES & $5.2k$ & $0.1k \,\, (1.19 \%)$ & $0.2k \, \, (2.38 \%)$ & $0$ & $8.4k$ & $\mathbf{3.92 \times}$\\
    8 CORES & $2.6k$  & $0.1k \,\, (2.27 \%)$ & $0.3k \,\, (6.81 \%)$ & $0$ & $4.4k$ & $\mathbf{7.52 \times}$\\
    \hline
    \hline
  %\bottomrule
\end{tabular}
\caption{The table shows the multicore execution profiling of the kernels. The measurements for multicore configurations are reported as an average of the measurements taken on each core. The percentage value highlights the impact of each measured contribution on the total execution cycles.}
 \label{tab:parallel_results}
\end{table*}
%\end{center}
\subsection{Multicore Execution Results}
\label{sec:multicore_results}
\begin{figure}[t]
    \centering
    \includegraphics[width= 0.9\linewidth]{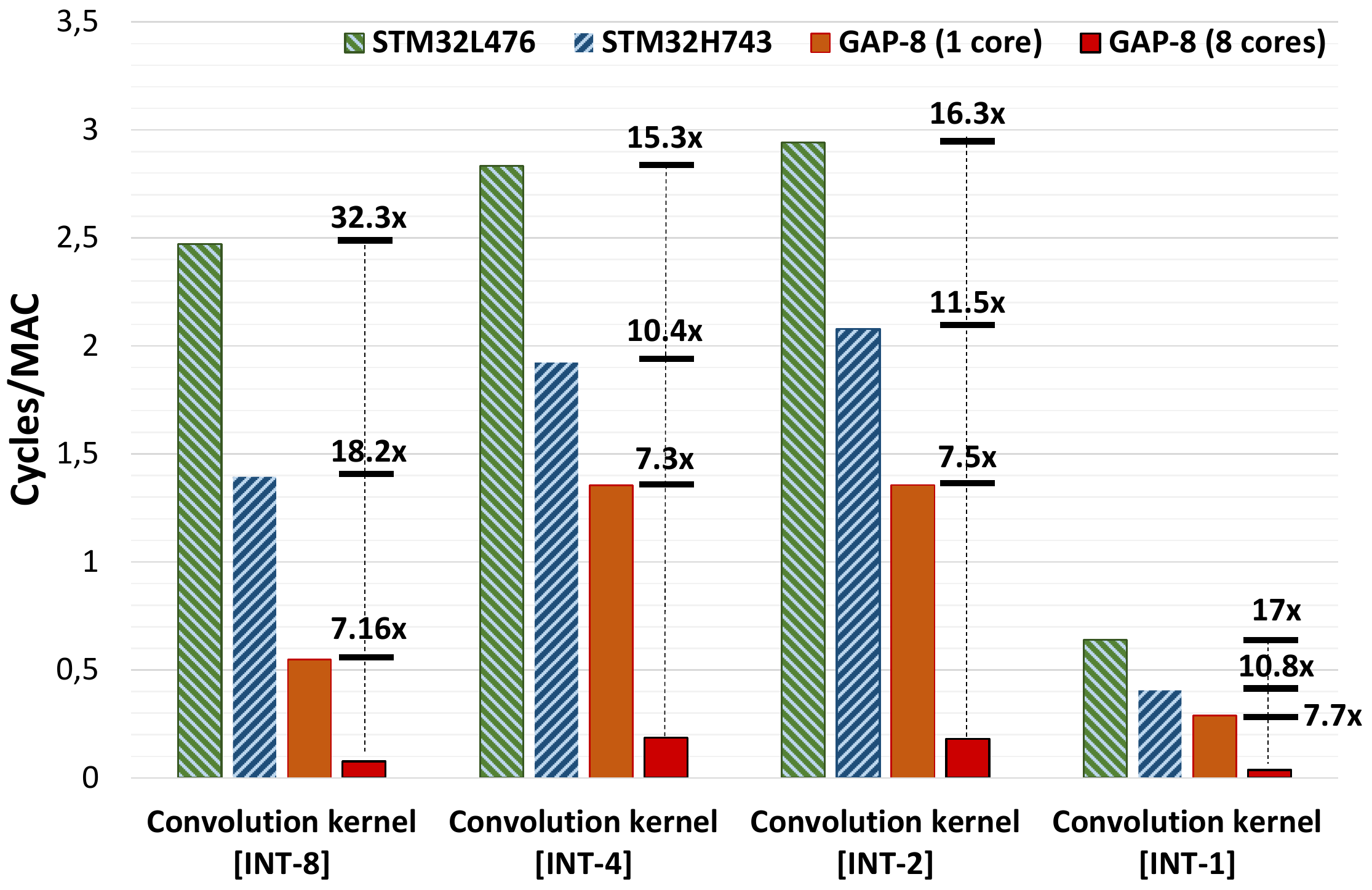}
    \caption{Comparison in terms of cycles/MAC between the PULP-NN conv kernels on one/eight core(s) of GAP-8 cluster and CMSIS-NN conv kernels on STM32L4 and STM32H7.}
    \label{fig:ARM_comp_multicore}
\end{figure}
In this section, we focus on the analysis of the multicore optimization of the kernels.
%
%In particular, it is important to understand well all the possible limits of the optimization proposed.
%
%For this reason, we provide for an exhaustive study of the performance we are able to achieve.
%
%To this purpose we fully exploit the \textit{performance counters} provided by the architecture.
%
Figure \ref{fig:ARM_comp_multicore} shows a comparison of the convolution kernels running on the 8-core cluster of GAP-8 with respect to the equivalent CMSIS-NN implementation on STM32H7 and STM32L4.
%
%The comparison with STM32H7 and STM32L4 is replicated also for the eight cores execution of the convolution kernels and reported in Figure \ref{fig:ARM_comp_multicore}, in terms of cycles/MAC.
%
It is possible to notice that, due to the additional operations required to execute sub-byte kernels, their overall cycles/MAC are 0.186 for INT-4 and 0.181 for INT-2, both 2.4$\times$ higher than the INT-8 case.%execution time is $2.4 \times$ higher than the INT-8 case.
%
%It is possible to note that, due to the additional pack/unpack operations required to execute sub-byte kernels, the overall execution time when running INT-4 and INT-2 convolution layers is 2.4x larger than the INT-8 case.
However, we can notice how the software-efficient exploitation of the parallel processors cluster provides almost linear speedups ($7.16 \times$ to $7.7 \times$) with respect to the single core configuration, leading to a dramatic improvement of performance with respect to the equivalent execution on sequential RV32IMC (where the overall speedup passes from 8.8$\times$ of the single-core execution to up 63$\times$ when considering 8-cores) and on single-core ARM architectures ($10 \times$ to $32 \times$). This huge performance gain enables the exploitation of the benefits of heavily quantized neural networks in terms of memory footprint, still performing one order of magnitude better than state-of-the-art ARM-based implementations.
%questo va migliorato
%

To provide more insight on the multi-core optimizations, we present an exhaustive study of the performance achieved on the parallel cluster of GAP-8.
First, we measure the amount of executed instructions per each core providing an indication of the Amdhal's limit of the kernels, i.e. the amount of cycles lost due to non-parallelizable code.
%, usually due to the presence of sequential sections in the code.
%
As a second point, we measure the the number of cycles in which the cores are not waiting on a barrier (active cycle).
%which indicates the ratio of cycles lost due to unbalanced parallelization.
%
Then we measure the architectural sources of overhead: number of cycles lost due to contention on the shared TCDM, cycles lost due to instruction cache stalls and cycle lost due to load stalls (read after write).
The results for the convolution and fully-connected kernels are summarized in Table \ref{tab:parallel_results}.
%All the parallelization results, for the convolution and the fully connected kernels, are summarized in Table \ref{tab:parallel_results}.
%

%

%What comes out from the analysis of the results is that we achieve essentially linear speedup when offloading the computation to either two or four cores of the GAP8 cluster.
%We achieve essentially linear speedup when offloading the computation to either two or four cores of the GAP8 cluster.
%
%The eight cores configuration instead, deserves to be discussed separately, to justify all the cycles lost.
%
Considering the convolution kernel, we achieve a Speedup of $7.16 \times$ with eight cores.
%
%Nevertheless it is interesting to understand what is the cause of the loss of cycle.
%
By analyzing the table we can notice that the Amdahl's limit of the kernels is around $8 \times$ (thus, ideal), but we lose a small number of cycles due to architectural overheads: the
%, approximately $32\;k$.
%The primary cause of loss is due to execution unbalance, where we loose around $32 k$ cycles.
%
$67\%$ of this overhead is due to I\$ non-idealities, $8\%$ is due to load stalls and $20\%$ is due to TCDM contention, which is reasonable as there are eight cores that access the same shared L1 memory.
The number of I\$ stalls increases with the number of cores due to the increasing contentions in the shared cache banks \cite{loi2018quest} (the banking factor of 8 can not completely remove the conflicts), on top of the I\$ misses due to the large inner loop of the kernel.
%
%After that we notice a discrepancy between the active cycles and the total cycles. Since these cycles are not captured by any measurments we did, we must conclude that they are due to noise or to some other cause.
%
%An additional cause of cycles loss could be the misaligned instructions in memory. These indeed would take two cycles to be decoded.
%
The parallel execution of the fully connected layer presents a speedup higher than the convolution kernel mainly thanks to the reduction of I\$ stalls due to the smaller size of the kernel. 
%
%This situation does not stress the instruction cache and what we notice is a reduced number of I\$ stalls.
%with a subsequent better results as regards the execution unbalance.
%
The speedup is never lower than $7 \times$ also when considering the max-pooling and ReLU kernels running on eight cores.
%A glance to the results of the Max pooling parallelization suggests that this is the case where we achieve the lower Speedup which however is not lower than $7 \times$.
%
%Repeating the same analysis of the convolution kernel would bring to the same explanation of the loss of cycles already highlighted before.
%
% We conclude that the approach we used to parallelize the kernels over the cores of GAP8 cluster reveals to be very efficient.
% %
% We are able to achieve a near linear speedup in all cases. 
%

%%% KERNEL EXPLORATION %%%
\subsection{Kernel Exploration}
\label{sec:kernel_expl_results}
\begin{figure}
  \centering
  \includegraphics[width=0.9\linewidth]{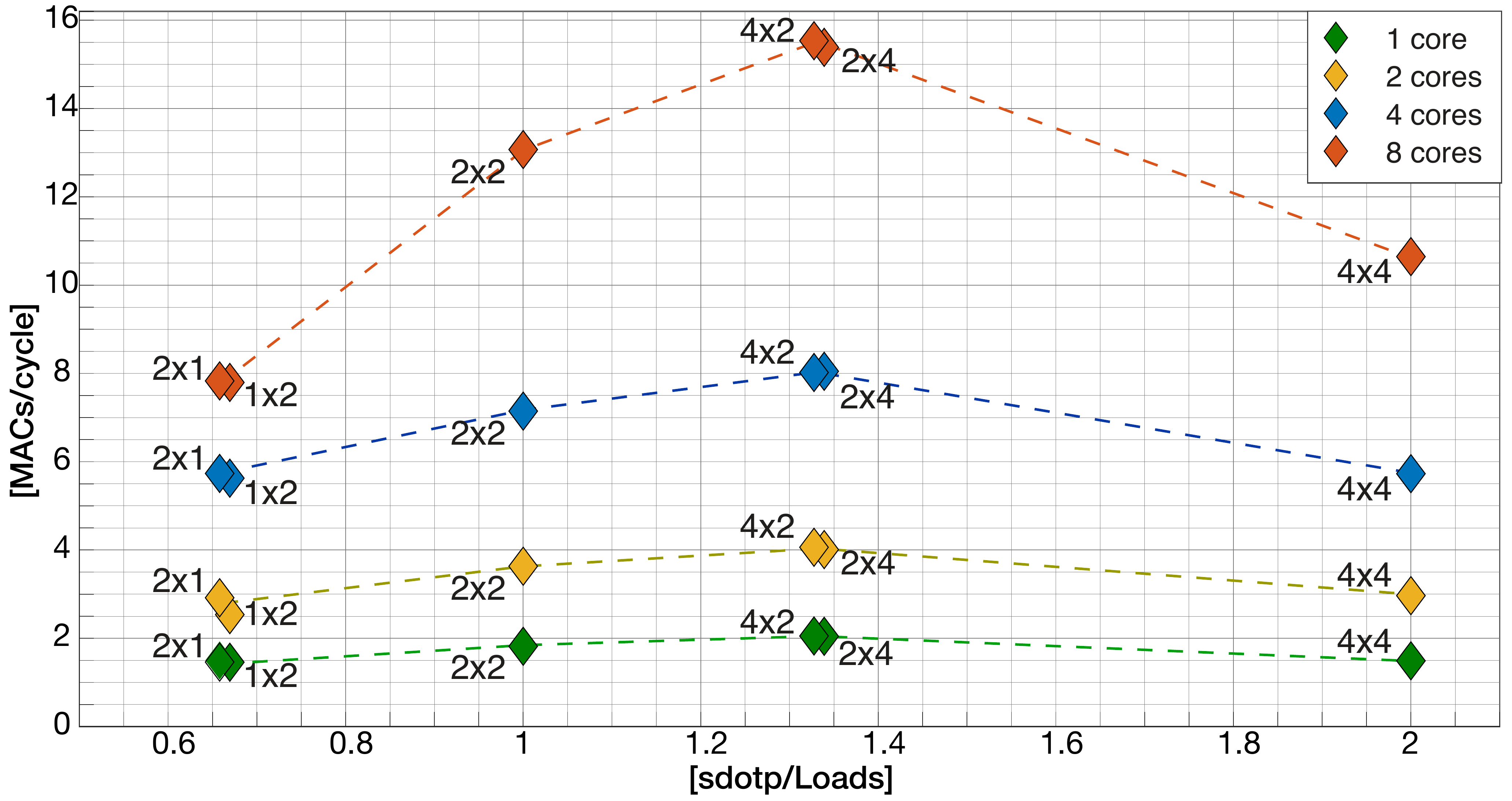}
  \caption{Performance of the convolution layer considering different sized matrix multiplication kernels. On the x-axis we show the sdotp to load ratio to clarify how many \textit{sdotp4} (equivalent to 4 MAC) we can set with one load. The label of each point of the graph, in the form of $a \times b$, specifies the kernel size considered. $a$ is the number of output features computed by the kernel, $b$ is the number of output activations.}
  \label{fig:kernel_exploration}
\end{figure}

The exploration of the matrix multiplication kernel size design space is carried out for the INT-8 operands, considering sizes ranging from 1$\times$2 to 4$\times$4.
%In this section, we highlight the experimental results of the matrix multiplication kernel size exploration.
%
%As introduced in the previous section, we explore different kernel sizes, to maximize the MAC to Load ratio, which directly measures the amount of data reuse.
%
%As this ratio grows, the throughput of the kernel, i.e., the number of MACs per cycle, grows as well.
%
%The analysis is carried out for INT-8 data types.
%
%Our exploration space consists of kernel sizes ranging from 1x2 to 4x4.
%
The results are summarized in Figure \ref{fig:kernel_exploration}.

%
%% LABEL...
%The figure shows the results of the matrix multiplication kernel sizes exploration. Along x-axis, we report the sdotp to loads ratio, and along y-axis, we report the throughput of the computation, measured as the number of MACs per cycle.
A peak throughput of 15.5 MACs/cycle is reached when we consider a convolution kernel with a 4$\times$2 sized matrix multiplication kernel running over eight cores of the cluster, achieving a result of just 1.01 LD/ST per MAC. This result translates in an overall efficiency of $49\%$ in terms of MAC utilization, only a factor of 2 from the theoretical peak achievable (32 MACs/cycle) on a cluster of eight programmable cores with SIMD MAC units, i.e. considering the MAC units constantly fed.
%
%We achieve to compute $15.5 MACs/cycle$, in this case.
%
Nearly the same throughput is achieved with the 2$\times$4 sized kernel, as the almost overlapping points in the graph suggests.
%, but reusing more data along the channels (4$\times$2) comes at a lower extra memory footprint to build the im2col buffers with respect to the 2$\times$4 sized kernel.}
%

Then, the optimal sized kernel has been chosen taking into account also the extra memory footprint needed to build the im2col buffers in the two configurations, which results to be lower for the 4$\times$2 solution (see section \ref{sec:size_explo} for more details). As regards the 1$\times$2, 2$\times$1 cases, they appear to be inefficient, as the amount of data reuse is meager and we pay the overhead due to the higher number of loads.
For these configurations, the MAC to load ratio is slightly higher than $1$.
The 4$\times$4 case instead would demonstrate to be the best, since the first indication of ideal data reuse is equal to 8 (MAC/load).
%
%Unfortunately we have to face with the architectural limitations.
%
However, to set a 4$\times$4 sized matrix multiplication kernel inner loop we should have at least 24 registers available (16 for the accumulators and 8 for the operands), while the target RISC-V, like most MCU-dedicated micro-architectures, has a register file with 32 general purpose registers.
With only eight usable registers, the compiler has to spill variables to the stack to make room for the accumulators and operands, leading to significant performance degradation.
%
% What comes out is that such a sized kernel runs out the available resources and this immediately leads to a performance degradation since the compiler has to rearrange the computation to fit the available registers.
%
%The kernel indeed run out the 32 registers available in the register file of the architecture considered, as well as the architecture of most of the off-the-shelf microcontrollers.
%
%

\subsection{Comparison with GAP8 Native Library}
\begin{figure}
  \centering
  \includegraphics[width=0.9\linewidth]{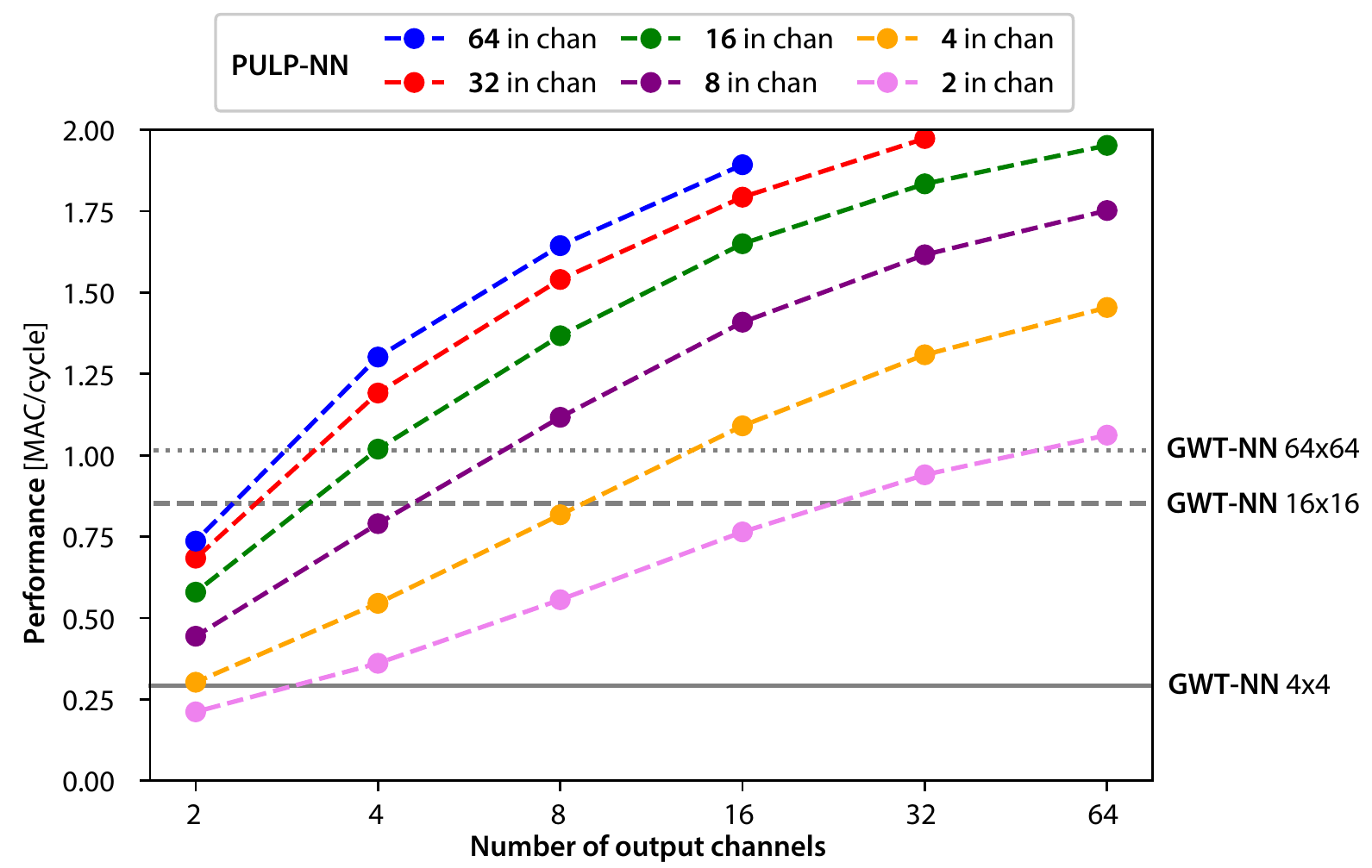}
  \caption{Comparison between PULP-NN using a 4$\times$2 kernel and the best result obtained by GWT-NN.}
  \label{fig:pulp_nn_vs_eric_nn}
\end{figure}

We compare our library with the optimized multi-core kernels that are openly distributed by GreenWaves Technologies as part of a proprietary tiling solution\footnote{https://github.com/greenwaves-technologies/autotiler.} and tailored for the GAP8 processor.
%The company producing the GAP8 processor, GreenWaves Technologies, also distributes a set of optimized multi-core kernels open-source as part of a proprietary tiling solution\footnote{https://github.com/greenwaves-technologies/autotiler.}.
We call this library GWT-NN.
In this section, we compare the performance of PULP-NN on INT-8 data with that provided by GWT-NN.
We focus on a 3$\times$3 kernel in terms of filter size as a representative example constituting the bulk of most SoA DNNs.

Differently from PULP-NN, GWT-NN operates spatially on CHW-formatted data with explicit convolution filters working in a sliding window fashion, and accumulation over an appropriately sized INT-32 buffer.
In the innermost loop, the GWT 3x3 kernel uses the register file to implement a sliding window and uses three \textit{sdotp4} instructions to implement a total of 9 multiply-accumulate operations.
\cite{gautschi2017near} and \cite{palossi201964mw} report further details with respect to this convolution kernel.

Figure~\ref{fig:pulp_nn_vs_eric_nn} shows a comparison between the two libraries when running on a single core of the GAP-8 cluster, in terms of performance in MAC/cycle.
For PULP-NN, the performance is swept by changing the number of input and output channels between 2 and 64 (only results from configurations fitting the L1 are shown).
We chose the biggest input spatial size (24x24) for which configurations with 64 input or output channels fit L1.
Conversely, for GWT-NN, performance is substantially independent of the number of in/out channels, but only on the spatial size of the input image; therefore, we fix their input/output channels at 4 and have them sweep their input size between 4, 16, and 64 pixels height/width.

As visible from Figure~\ref{fig:pulp_nn_vs_eric_nn}, PULP-NN outperforms GWT-NN for all small images, and in most cases of spatially bigger images by a significant margin.
This is due to a combination of two effects: the 3x3 sliding window requires three loads and three \textit{sdotp4} per output pixel, yielding a lower \textit{sdotp4} per load ratio (1) with respect to the 4$\times$2 PULP-NN kernel (1.4); moreover, only three MAC are used per each \textit{sdotp4}, yielding a further loss of 25\% in terms of efficiency.
Consequently, the GWT-NN kernel is mostly competitive when the spatial size of the feature maps is much higher than the number of channels, e.g., in the first layer of a CNN. While, when the number of input/output channels is high, which typically represents the majority of the workload for state-of-the-art deep networks topologies \cite{howard2017mobilenets}, PULP-NN can achieve as much as a +89\% speedup with respect to GWT-NN.

\subsection{Comparison with State-of-the-Art Architectures}
\label{subsec:SoAcomparison}
\begin{figure*}
    \centering
    \includegraphics[width=\linewidth]{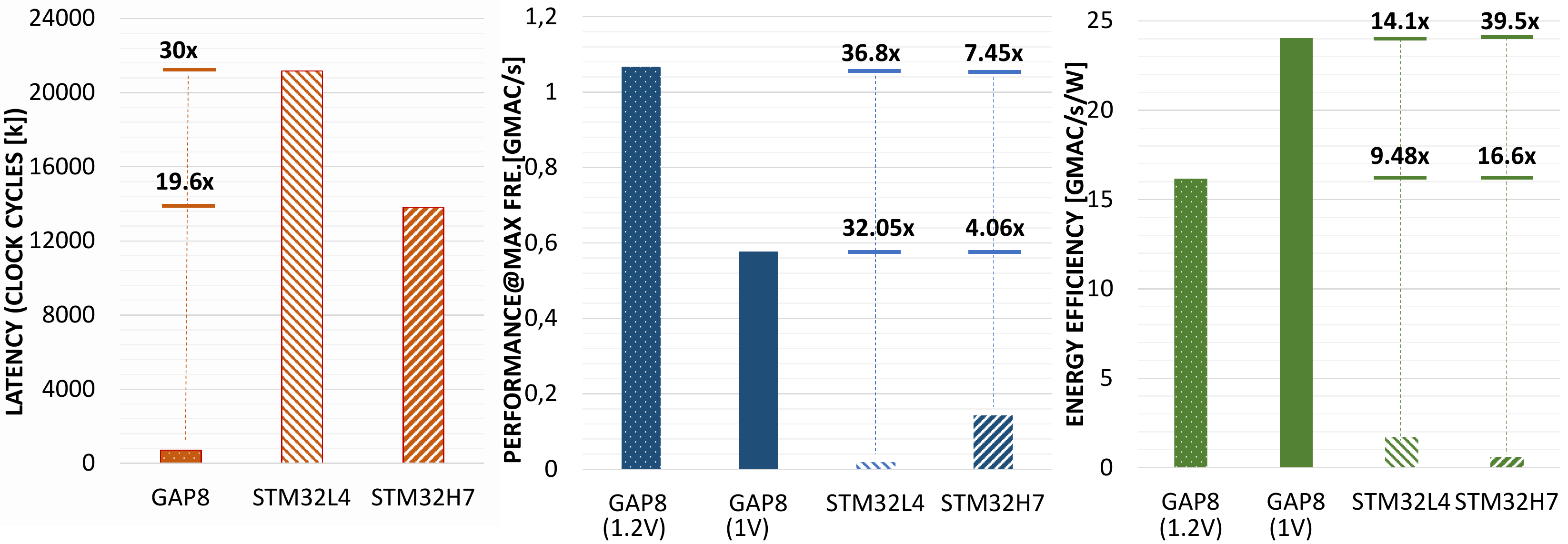} 
    \caption{This figure shows the execution cycles, the performance (at the maximum frequency) and energy efficiency (at the lowest consumption configuration) to infer the entire QNN on GAP8, STM32L4 and STM32H7 microcontrollers.}
    \label{fig:en_eff&perfo}
\end{figure*}

To assess the library performance on an inference task, we run a full QNN, trained on CIFAR-10 dataset, on GAP-8, using PULP-NN back-end library. For comparison purposes, we run the same network also on State-of-the-Art edge of IoT ARM Cortex-M based microcontrollers (STM32H7 and STM32L4), using CMSIS-NN.
STM32H7 and STM32L4 were chosen as representative of popular high-end and low-end MCU systems, with a clear trade-off between performance and energy efficiency. The comparison with these two popular computing platforms allows to analyze where our results lay in terms of trade-off between computing performance and energy efficiency.
%\textcolor{blue}{The direct comparison with high-end (STM32H7) and low-end (STM32L4) edge MCUs allows us to analyze where our achieved results lay in terms of trade-off between computing performance and energy efficiency.}
%
%CIFAR-10 is an open-access dataset for image classification purposes. It consists of 60,000 color images, 32x32 sized which are divided into 10 output classes.
%
The implemented network topology is composed by three convolution layers and one fully-connected layer, consisting of $26.7\,k$ parameters and $6.56$ MMACs in total\footnote{The layer parameters can be found at: https://github.com/ARM-software/ML-examples/tree/master/cmsisnn-cifar10}.
The weights and the activations are quantized to INT-8 format.
Such a topology is already used on IoT edge devices (MCUs) and also used by ARM to validate Neural Networks on low-power microcontrollers such as STM32L4 or STM32H7.

On GAP-8, the RGB image is initially stored in the L2 memory and brought in the L1 memory before the start of the inference task, through a DMA transfer.
The activation values are then kept in the L1 memory to save on memory transfer overhead. Before the execution of each convolution or linear kernel the weights, initially residing on  L2  memory, are brought in  L1 through DMA as well. Also the im2col buffers are kept in L1 memory.
On the STM32L4 microcontroller, the entire network is stored in the first level of memory, which consists of 128 kB SRAM. On STM32H7 the network is stored in SRAM as well and we enable also the harware data cache which is provided by the MCU architecture.

%Featuring the STM32L4 microcontroller only one level of SRAM (160 kB in size), the entire network has been stored there.
%Contrary on GAP-8  and  STM32L4, which does not feature a hardware data cache, on the  STM32H7 such a memory has been kept enabled during the inference.

In the single core configuration, we are able to infer the entire network in 28.6 ms, when GAP-8 runs at 170 MHz.
%$4.87\times10^6$ cycles.
%
We achieve almost linear speedup when considering two and four cores, $1.99 \times$ and $3.79 \times$ respectively.
With eight cores the speedup is slightly less than $7 \times$.
%
%We run the same network on an  \cite{STM32L476} boards using CMSIS-NN, and we compare the result with our solution.
%
Figure~\ref{fig:en_eff&perfo} shows the comparison of PULP-NN implementation of the network on GAP-8 with respect to the CMSIS-NN implementation on STM32H743 and STM32L467 in terms of execution cycles, performance (i.e. also considering the maximum operating frequency of the devices), and energy efficiency.
%

%
%The execution time to infer the QNN on the STM32H7 board is $13.73\times 10^6$ cycles, increasing to $21.16\times 10^6$ on the STM32L4 due to its less advanced internal architecture.
%
%The latter is the second baseline to compare our solution and the results, in terms of speedup, complete the Table \ref{tab:cifar10ex}.

%In terms of execution cycles, the inference of the QNN, using PULP-NN on a fully exploited GAP8 cluster, results in a speedup of $19.5 \times$ with respect to the inference on an ARM Cortex-M7 based processor using CMSIS-NN.
%
%Figure \ref{fig:en_eff&perfo} shows the performance, and the energy efficiency achieved running the entire CIFAR-10. 
Our PULP-NN CIFAR-10 achieves a peak performance of 1.07 GMAC/s at the frequency of $170$ MHz and the supply voltage of 1.2 V on GAP-8, inferring $241$ frame per second (fps) with an energy per inference of $0.27$ mJ/frame. 
The performance is $7.45 \times$ better than the STM32H7 and $36.8 \times$ better than the STM32L4. The energy efficiency achieved at this operating point is $16.1$ GMAC/s/W, $16.6 \times$ higher than the STM32H7 and $9.48 \times$ higher than STM32L4. 
At the same time, at the best energy point, at the supply voltage of 1V, PULP-NN achieves a performance of $577$ MMAC/s on GAP-8, with energy efficiency of $24$ GMAC/s/W, inferring $127$ fps with $0.19$ mJ/frame, and outperforming STM32H7 by $4.06 \times$ and STM32L4 by $32.05 \times$ in terms of performance and by $39.5 \times$ and $14.1 \times$ the same devices respectively, in terms of energy efficiency.
%At the same time, at the best energy point, at the supply voltage of 1V, PULP-NN achieves a performance of $577 MMAC/s$ on GAP-8, with energy efficiency of $24 GMAC/s/W$, inferring $127\,fps$ with $0.19\,mJ/frame$, and outperforming STM32H7 by $39.5 \times$ and $4.06 \times$ and STM32L4 by $14.1 \times$ and $32.05 \times$ in terms of energy efficiency and performance, respectively.
%
%In the same terms, we outperform also STM32L4 by $36.8 \times$ and by , respectively.
%

\subsection{Discussion}

In this work we demonstrate that coupling optimized software libraries with a parallel ultra low power computing platform we achieve energy proportionality where, as opposed to commercial ARM-based solutions, we do not have to trade performance with energy efficiency, paving the way to fully software programmable CNN inference at the edge of the IoT.
However, sub-byte kernels still suffer from drop-off in performance when compared to the INT-8 ones, despite their execution on GAP-8 performs more than one order of magnitude better with respect to MCU-based SoA solutions. The overhead, as highlighted in section \ref{sec:RISC-V_impl_res}, is due to the hardware support of the target architecture only for 8-bit SIMD instructions, which makes necessary to introduce additional packing and unpacking functions.
The sub-byte precision QNNs though, provide several advantages when deployed at the edge, since their memory footprint decreases linearly with the bit-width used to represent weights and activations \cite{hubara2017quantized}, making them more suitable to fit the limited memory capacity of MCU-like devices. Moreover, it has the potential to increase the energy efficiency, crucial for battery-powered devices \cite{moons2017minimum}. Recent research demonstrated that, by exploiting specific retraining techniques, the accuracy drop can be kept under control, leading to a cumulative loss which is acceptable for many IoT applications \cite{rusci2019memory}. Hence the research community is focusing more and more on the study and implementation of strongly quantized NNs. It is therefore important going further in the work presented in this paper to exploit fully the potential of heavily quantized networks on fully programmable edge devices. From the hardware perspective, providing the target ISA with sub-byte hardware SIMD operations will be a step forward to eliminate the software overhead and to double, at least, the performance and the energy efficiency with respect to the current optimal 8-bit solution.
\sloppypar

\section{Conclusion}
\label{sec:Conclusion}
We have presented PULP-NN: an optimized library to run QNNs at the edge, targeting INT-8, INT-4, INT-2, and INT-1 data operands.
We showed that, by optimizing the library with the SIMD extensions and bit manipulation instructions of the targeted architecture, we heavily increase the performance of each kernel by up to 63x with respect to a corresponding RISC-V \textit{IMC} implementation, in an eight core cluster configuration.
%
%We presented an exploration of the matrix multiplication kernel size design space, and we found the best data reuse condition in a 4x2 sized kernel, achieving, in this case, a peak performance of $15.5 \; MAC/cycle$.
%
%Using such a sized kernel in convolution layer allows reducing its execution latency by 20 \%.
%
Running an entire INT-8 QNN on GAP8 showed us that we can achieve a speedup (in terms of cycles) of $19.49 \times$ with respect to the inference of the network on an STM32H7 microcontroller, using CMSIS-NN library.
Furthermore, the energy efficiency achieved on GAP8 results to be $24\; GMAC/s/W$, $14.1 \times$ higher with respect to the one obtained with STM32L4 board.
We conclude the same also for the performance: GAP8 achieves $1.066 \;GMAC/s$, which is $7.45 \times$ higher than the performance of STM32H7 board.
%

%

%\textcolor{blue}{From the hardware perspective, providing the target ISA with sub-byte SIMD operations will be a step forward not only to eliminate the software overhead but to unlock the full potential of strongly quantized NNs at the edge of IoT in terms of reduction of memory footprint and operation efficiency, at least doubling the performance and the energy efficiency with respect to the current optimal 8-bit solution (presented in this paper), on fully programmable devices.}
%
%\textcolor{blue}{From the software perspective, our future work will focus on the exploitation of the DMA to build an automatic tiling solution for hiding the latency of moving intra-layer data from/to the off-cluster L2 memory, when running a QNN whose layer parameters do not fit the L1 memory. These solutions will form a complete and extremely flexible tool to ease the deployment of QNNs at the edge for IoT applications.}

%\textcolor{blue}{A final comment on sub-byte kernel performance. As seen before, the sub-byte kernels suffer from significant drop-off in performance due to the Xpulp ISA support only for 8-bit SIMD operation, making necessary additional packing and unpacking instructions to run INT-4 or INT-2 kernels. Enabling, in future, sub-byte hardware SIMD operations will not only eliminate such additional software overhead but unlock fully the sub-byte QNN potential, in terms of memory footprint and operational efficiency, doubling or even quadrubling  the performance with respect to the current optimal 8-bit solution, on fully programmable edge devices.}

\section*{Acknowledgements}

This work was supported in part the OPRECOMP (Open trans-PREcision COMPuting) project founded from the European Union's Horizon 2020 research and innovation program under Grant Agreement No. 732631.

% The next two lines define the bibliography style to be used, and the bibliography file.
\bibliographystyle{IEEEtran}
%\bibliography{sample-base}

\end{document}